\pdfoutput=1

\documentclass[10pt,twocolumn,letterpaper]{article}

\usepackage[pagenumbers]{cvpr} %

\newcommand{\supp}{App.}
\newcommand{\abslinks}{\par\smallskip\noindent\textbf{Project page:} \href{https://uservan.github.io/GoldiBench}{uservan.github.io/GoldiBench}\\
  \textbf{Code:} \href{https://github.com/uservan/GoldiBench}{GitHub}\quad
  \textbf{Data:} \href{https://huggingface.co/datasets/VanWang/GoldiBench}{Hugging Face}}

\providecommand{\supp}{Supp.}
\providecommand{\abslinks}{}

\makeatletter
\renewcommand\paragraph{\@startsection{paragraph}{4}{\z@}{2.4ex \@plus .3ex \@minus .2ex}{-1em}{\normalfont\normalsize\bfseries}}
\makeatother

\usepackage{colortbl}
\usepackage{multirow}
\definecolor{CliffBlue}{HTML}{1F4E9C}

\newcommand{\bench}{GoldiBench\xspace}

\usepackage{stfloats}

\usepackage{placeins}

\definecolor{cvprblue}{rgb}{0.21,0.49,0.74}
\usepackage[pagebackref,breaklinks,colorlinks,allcolors=cvprblue]{hyperref}

\def\confName{CVPR}
\def\confYear{2027}

\title{GoldiBench: Right Tools, Wrong Amount\\Benchmarking Agents on Effect Judgement in Professional Editing}

\author{Wang Yang$^{1}$ \quad Xinpeng Li$^{1}$ \quad Yanyan Zhang$^{1}$ \quad Chaoda Song$^{1}$\\
Shouren Wang$^{1}$ \quad Debargha Ganguly$^{1}$ \quad Vikash Singh$^{1}$ \quad Chuang Ma$^{2}$\\
Xiaoran Cai$^{3}$ \quad Kai Ye$^{1}$ \quad Xiaotian Han$^{1}$ \quad Vipin Chaudhary$^{1}$\\[4pt]
$^{1}$Case Western Reserve University \quad $^{2}$NII, Japan \quad $^{3}$Columbia University
}

\begin{document}
\maketitle
\begin{figure*}[b]
  \centering
  \includegraphics[width=\textwidth]{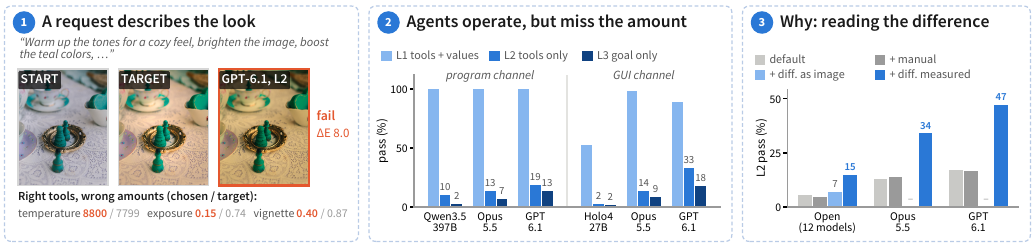}
  \caption{\textbf{Agents operate editing software, but do not reach the requested effect.}
  (1)~A request in \bench describes a look; the hidden program behind TARGET fixes how much of each tool it takes.
  Given only the tools (L2), GPT-6.1 picks the right ones but the wrong amounts and stops after one render.
  (2)~Pass rate falls from L1 (tools and values given) to L2 (tools only) and L3 (goal only), on programs and in the GUI
  (mean over five applications).
  (3)~At L2, a manual or the difference shown as an image does not help; the measured TARGET$-$START difference does
  (Sec.~\ref{sec:why}).}
  \label{fig:teaser}
\end{figure*}

\begin{abstract}
Real editing requests describe a look, not the steps to reach it, so an agent that serves them must judge how much of
each tool the look takes. Existing benchmarks do not measure this: their tasks either state the tools and values, or
leave the amount open and are scored by VLM judges that pass more than half of half- and triple-strength results.
We introduce \bench, a benchmark of 500 editing tasks in five professional applications (darktable, RawTherapee, GIMP,
Inkscape, Blender), solvable by writing a program or by operating the GUI.
Each task pairs a START image with a TARGET rendered from a hidden program, so the amount is fixed by a known target; the
agent's edit is re-rendered and compared with TARGET, and the same task is posed with the tools and values, the tools
only, or the goal only, separating operation from effect judgement.
Across 25 open, GUI and closed models, agents operate well (99--100\% with programs from 4B up when told the values) but fall off a cliff
when they must judge the amount: given only the tools, the best model passes 19\% of the tasks with programs and 33\%
through the GUI, and scale does not close the gap.
Manuals barely help; the measured TARGET$-$START difference more than doubles the pass rate, but the same difference
shown as an image does not for open models. Allowed four renders, agents almost always stop after the first; forced to compare, the
strong models double their pass rate. Agents know the tools, but neither read nor check the amount.\abslinks
\end{abstract}

\section{Introduction}
\label{sec:intro}
Computer-use agents can now operate professional software: they open dialogs, set fields and write scripts for
image editors and 3D tools~\cite{osworld,opencua,uitars,cogagent,seeclick}.
Real editing requests, however, rarely say what to operate.
They describe a look, ``warm it up for a cozy feel, brighten it, darken the corners'', and leave the tools and the
amounts to the editor: in two request communities, 99--100\% of the edit requests name no tool (Sec.~\ref{sec:mot-real}).
Serving them takes two skills, \textbf{operation}, carrying out a given edit in the software, and \textbf{effect
judgement}, deciding which tools and how much of each reach the requested look.

\textbf{Existing benchmarks cannot measure effect judgement.}
Tasks that state the tools and values (57\% of the PSBench~\cite{psbench} easy tasks) can be solved by entering
them; on OSWorld's GIMP tasks, agents succeed on 85\% of such operation tasks but on 39\% of the edits that name no
tool (Sec.~\ref{sec:mot-explicit}).
Tasks that leave the amount open have no determinate answer and are scored by a VLM judge, which cannot tell the
amount: it passes 55\% of half-strength and 58\% of triple-strength results, against 84\% of correct ones
(Sec.~\ref{sec:mot-vague}).
Whether agents can judge effects is unknown.

\textbf{\bench.}
We build a benchmark that measures effect judgement directly.
Each task is a START image and a TARGET rendered from a hidden program of real tool settings, so ``how much'' is fixed
by a known target.
The agent's edit is read back from the project file, re-rendered and compared with TARGET (CIEDE2000 and closeness),
with no judge in the loop.
The same task is posed at three levels that withhold more and more of the program: tools and values (L1), tools
only (L2) and the goal only (L3), so a failure can be attributed to operation or to judgement.
\bench has 500 tasks in five open-source applications in four domains (photo, raster, vector, 3D), each
solvable by writing a program or by operating the GUI, with one evaluator (Fig.~\ref{fig:teaser}).

\begin{figure*}[t]
  \centering
  \includegraphics[width=\textwidth]{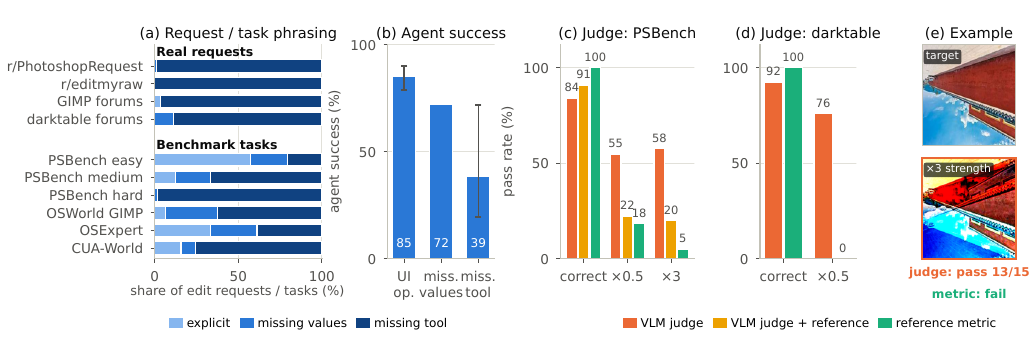}
  \caption{\textbf{Benchmarks test operation but cannot verify effect judgement.}
  (a)~Real requests name no tool (960 posts), while many benchmark tasks state the tool and its values.
  (b)~Published agents succeed on UI-operation tasks far more than on edits that name no tool
  (OSWorld GIMP, 94 agents, 95\% CI).
  (c)~The PSBench judge passes half- and triple-strength results about as often as correct ones;
  showing it the reference helps but does not close the gap to a reference metric (CIEDE2000 $\le 3$).
  (d)~The same holds on real darktable renders; our metric fails every half-strength render.
  (e)~A PSBench task (``flip the image vertically''): at triple strength the colours are destroyed, yet 13 of 15
  judgements pass it.
  Full results in \supp~\ref{app:motivation}.}
  \label{fig:motivation}
\end{figure*}

\textbf{Findings.}
We evaluate 25 models: twelve open VLMs (0.8B to 397B), eleven GUI agents and two closed ones.
\begin{itemize}[leftmargin=*,topsep=2pt,itemsep=1pt]
  \item \emph{Agents operate, but miss the amount.} Given the tools and values, every program model from 4B up passes
        99--100\%; given only the tools, the best passes 19\%; given only the goal, 13\%
        (GUI: 33\% and 18\%).
        The GUI shows the same cliff (Opus 5.5: L1 99\%, L2 14\%), and model size raises L1 far more than L2 or L3.
  \item \emph{The bottleneck is reading the difference, not knowing the tools.} Manuals and tool references help
        little (at most $+$8 points, GPT-6.1 only). Telling the model the measured TARGET$-$START difference more than doubles L2 (GPT-6.1 17\% $\to$ 47\%),
        while showing the same difference as an image does not (open models).
  \item \emph{Agents do not check their result.} Allowed four renders, the closed models stop after the first in
        99.8--100\% of the L2 episodes; forced to compare, they more than double their L2 pass rate.
\end{itemize}

\textbf{Contributions.}
(i) \bench, a benchmark of determinate editing tasks across five applications and two interaction channels that
separates operation from effect judgement, with a reference metric that verifies the amount against a known target,
which VLM judges cannot do reliably;
(ii) an evaluation of 25 open, GUI and closed models that locates where they fail;
(iii) controlled interventions showing that the gap is in perceiving and checking the effect, which tells
agent builders what to fix.

\section{Motivation: Benchmarks Test Operation but Cannot Verify Effect Judgement}
\label{sec:motivation}

We show that existing benchmarks measure operation but cannot verify effect judgement
(Fig.~\ref{fig:motivation}; setups in \supp~\ref{app:motivation}).

\subsection{Professional use requires effect judgement}
\label{sec:mot-real}

\textbf{Real requests.}
We sampled 960 public posts from four editing communities (r/PhotoshopRequest~\cite{realedit}, r/editmyraw, and the GIMP and darktable
forums) and had two LLMs label each edit request as \emph{explicit} (tool and values named),
\emph{missing values}, or \emph{missing tool} (\supp~\ref{app:r1}).
In the two request communities, 99\% and 100\% of the requests name no tool at all, and none names both a tool and
its values (Fig.~\ref{fig:motivation}a, top).
An agent serving them must choose tools and amounts, not only operate software.

\subsection{Explicit tasks test operation only}
\label{sec:mot-explicit}

\textbf{Task phrasing.}
Applying the same rule to the tasks of existing benchmarks shows that many state the answer:
57\% of the PSBench~\cite{psbench} easy tasks are explicit, and 16--33\% of the OSExpert~\cite{osexpert} and
CUA-World~\cite{cuaworld} editing tasks (Fig.~\ref{fig:motivation}a, bottom; \supp~\ref{app:e1}).

\textbf{Agent success by task class.}
Joining these classes with the published OSWorld-Verified results on the OSWorld~\cite{osworld} GIMP tasks
(94 agents), agents succeed on 85\% of the UI-operation tasks but on only 39\% of the edits that name no tool
(Fig.~\ref{fig:motivation}b; \supp~\ref{app:e1b}).
PSBench reports the same drop, from 69.9\% on easy to 0.0\% on hard tasks (without layers).
Explicit tasks mainly measure operation; as task classes also differ otherwise, we compare levels on the same
task (iii).

\begin{figure*}[t]
  \centering
  \includegraphics[width=\textwidth]{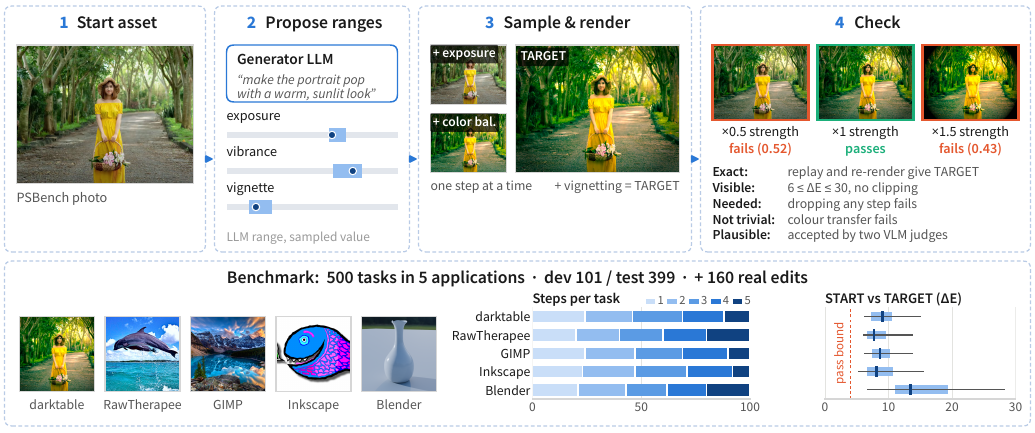}
  \caption{\textbf{Data pipeline and the resulting benchmark.}
  Top, on one darktable task: (1) a start asset; (2) a generator LLM proposes an intent and a value range for every
  field it sets (bar: the library range, shaded: the LLM's range, dot: the sampled value); (3) the program is rendered
  step by step into TARGET; (4) the task is kept only if it passes every check, \eg the program at half or 1.5 times
  its strength must fail (closeness in brackets).
  Bottom: one target per application, tasks by program length, and the CIEDE2000 between START and TARGET
  (box: quartiles, line: median, whiskers: 5th--95th percentile; dashed: the pass bound $\Delta E = 4$).}
  \label{fig:pipeline}
\end{figure*}

\subsection{Vague tasks need effect judgement but cannot be verified}
\label{sec:mot-vague}

Benchmarks score indirectly without a determinate answer.

\textbf{Evaluator audit.}
In the released evaluators (\supp~\ref{app:e2}), PSBench's VLM judge never sees the reference,
OSWorld's ``increase/decrease'' checks test only the direction,
and 16 of 20 OSExpert GIMP evaluators are unreleased.

\textbf{Judge stress test.}
We gave the PSBench judge, with five open VLMs, PSBench's reference result and controlled wrong versions of it on 60 tasks
(\supp~\ref{app:e3}).
It passes 55\% of half-strength and 58\% of triple-strength results, against 84\% of correct ones;
a plain CIEDE2000 threshold against the reference passes 18\% and 5\% (Fig.~\ref{fig:motivation}c).
The passed results include visibly broken ones (Fig.~\ref{fig:motivation}e).
Showing the judge the reference reduces but does not remove the problem (22\% and 20\%),
and the outcome depends on which VLM is the judge.

\textbf{Judge stress test on darktable.}
The failure is not specific to synthetic candidates.
On darktable, with real renders of the hidden edit at half strength, the judge passes 76\% of them (Fig.~\ref{fig:motivation}d);
with the exact values stated in the instruction, GLM-5.3 passes them as often as correct ones (98\% vs.\ 98\%).
Our metric (Sec.~\ref{sec:metric}), which requires most of the START--TARGET distance covered, fails all.

\textbf{Editing without the software.}
An open image-editing model (Qwen-Image-Edit-2511) that never opens Photoshop passes 86--96\% of the hard PSBench
tasks under the PSBench judge, close to PSBench's reference results (93--97\%; \supp~\ref{app:c1}),
although its outputs are re-drawn pictures rather than edits.
A reference-free judge cannot tell an edit made in the software from a new picture that looks right.

\subsection{Design requirements}
\label{sec:mot-req}

Measuring effect judgement therefore needs three properties that no existing benchmark combines:
\begin{enumerate*}[label=(\roman*)]
  \item a \textbf{determinate target}, a start image and a target render, so that ``how much'' has an answer (Sec.~\ref{sec:mot-vague});
  \item \textbf{programmatic scoring}: the edit is re-rendered and compared with the target, not judged (Sec.~\ref{sec:mot-vague});
  \item \textbf{operation separated from judgement}: the same task is posed with less given, so that a failure
        can be attributed to one or the other
        (Sec.~\ref{sec:mot-explicit}).
\end{enumerate*}

\begin{figure*}[t]
  \centering
  \includegraphics[width=\textwidth]{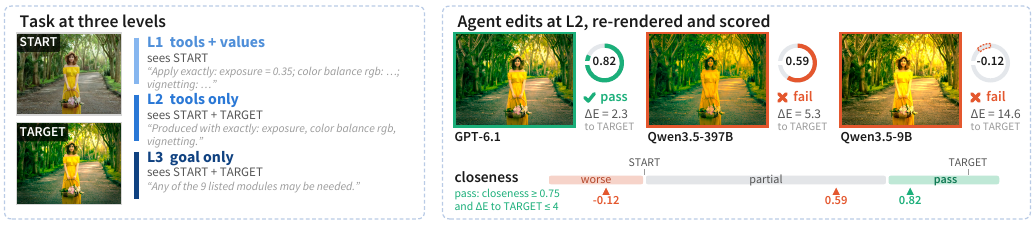}
  \caption{\textbf{A task and how it is judged} (the task of Fig.~\ref{fig:pipeline}).
  Left: what the agent is given at each level.
  Right: three agent edits at L2 (program channel, in the intervention of Sec.~\ref{sec:why} that reports the distance
  to TARGET after each round; standard runs report no distance),
  re-rendered from their programs; rings and the scale below show closeness (pass from 0.75). Qwen3.5-397B sets the exposure exactly but over-saturates and leaves too little
  contrast; Qwen3.5-9B moves away from the target (negative closeness).}
  \label{fig:task-example}
\end{figure*}

\section{Benchmark}
\label{sec:benchmark}

\bench realizes properties of Sec.~\ref{sec:mot-req} in five open-source applications, darktable and
RawTherapee (photo), GIMP (raster), Inkscape (vector) and Blender (3D), behind one OSWorld-format~\cite{osworld}
interface with a small adapter per application, so the evaluator is shared (\supp~\ref{app:apps}).

\subsection{Data pipeline}
\label{sec:pipeline}

\paragraph{Generated tasks.}
For each application we define an operation library of 7--11 common operations, with fields and ranges
(\supp~\ref{app:ops}).
Given a start asset and the library, a generator LLM writes an editing intent and a program of 1--5 steps with an
interval for every value; we sample values inside these intervals so that tasks do not collapse onto typical settings, and render the program to obtain TARGET (Fig.~\ref{fig:pipeline}).
Two generators (Qwen3.5-397B~\cite{qwen35}, GLM-5.3~\cite{glm5}) alternate; neither model does better on its own
tasks than other models do (\supp~\ref{app:val-tasks}).

\paragraph{Checks.}
\label{sec:checks}
A candidate becomes a task only if it is
\textbf{exact} (replay reproduces TARGET, $\Delta E \le 0.05$),
\textbf{visible} (START and TARGET clearly differ, without clipping),
\textbf{needed} in every step (removing any step fails the pass rule),
\textbf{not trivial} (a mean-colour shift or a colour transfer fails),
\textbf{sensitive} (the program passes at $\times0.9$ and $\times1.1$ strength and fails at $\times0.5$ and
$\times1.5$, so a task rewards the right amount, not only the right direction), and
\textbf{plausible} (both generators judge TARGET a result a professional would deliver; VLMs filter targets but
never score agents).
Of 7{,}093 candidates, 619 (9\%) pass, most rejections being an unneeded step (\supp~\ref{app:drops}).
We freeze 100 tasks per application and split them, stratified by program length, into public dev and held-out
test splits (about 20 and 80 per application).

\paragraph{Real tasks.}
To test whether generated tasks are realistic, we collect real edits with a recoverable program: on the
PlayRaw forum, users share a raw file with their darktable edit as an XMP sidecar, which \emph{is} the program.
160 of 312 replay exactly and pass our checks (results in Sec.~\ref{sec:val-tasks}).

\subsection{Task format and evaluation}
\label{sec:task-def}

\paragraph{Task and levels.}
A task is a start asset, a hidden program of the application's own operations with exact values, and an information
level; rendering the program gives TARGET, so every task has a known target; any legal edit (below) close enough to
it passes, whatever its settings.
The same task is posed at three levels (Fig.~\ref{fig:task-example}, left):
\textbf{L1} gives the tools with their values (no TARGET), so it measures \emph{operation};
\textbf{L2} gives TARGET and the tools without values, so the agent must judge how much of each to apply;
\textbf{L3} gives only TARGET, so the agent must also choose the tools.
Program, target and scoring are identical across levels, and the instructions are generated from the program by a
template (\supp~\ref{app:levels}), so a drop from L1 to L2 or L3 is caused by the withheld information alone.

\paragraph{Channels.}
In the \textbf{program channel} the agent writes the program as a list of tool settings~\cite{visprog,vipergpt} and, for up to four rounds, sees only its
render (no distance or score) and may stop early; in the \textbf{GUI channel} it operates the real application through screenshots, mouse and keyboard.
Both end in the same edit state and are scored alike.

\paragraph{Evaluation.}
\label{sec:metric}
The evaluator reads the edit state from the project, re-renders it with the application's own renderer, and compares
the render $R$ with the target $T$ in CIEDE2000~\cite{ciede2000} ($\Delta E$), relative to the start $S$:
\begin{equation}
  \text{closeness} = 1 - \frac{\Delta E(R,T)}{\Delta E(S,T)} .
\end{equation}
Closeness is 1 for an exact match, 0 for doing nothing and negative when the result moves away from the target.
A task passes if closeness $\ge 0.75$ and $\Delta E(R,T) \le 4$: most of the edit is done, and the remaining error
is small in absolute terms.
A legality gate fails results that use tools outside those given (the library at L3), change the geometry or import pixels, and
because the result is re-rendered from the edit state, pasting the target or editing pixels outside the application
cannot pass.
We report mean closeness and pass rates under other thresholds alongside (Sec.~\ref{sec:experiments}) and validate the
metric in Sec.~\ref{sec:validation}.

\subsection{Dataset statistics}
\label{sec:stats}

The benchmark has 500 generated tasks (dev 101, test 399) from 417 distinct assets and 160 real tasks
(Fig.~\ref{fig:pipeline}, bottom; Tab.~\ref{tab:bench-stats}).
Program lengths are balanced from one to five steps, and the median START--TARGET distance is 7.7--13.4 $\Delta E$
per application, well above the pass bound, so doing nothing or a small step in the right direction does not pass.
The GUI channel covers the tasks reproducible through the interface (53--100 per application and level).

\begin{figure}[t]
  \centering
  \includegraphics[width=\linewidth]{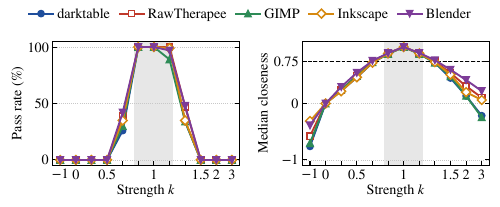}
  \caption{\textbf{Pass rate and median closeness versus edit strength.} Each numeric field of the hidden program is
  set to $k$ times its distance from the neutral value ($k=-1$ reverses the program); 100 tasks per application.
  Shaded: strengths the checks require to pass; dashed: closeness threshold.}
  \label{fig:strength}
\end{figure}

\section{Validation}
\label{sec:validation}

We validate the metric, the tasks and the measurement in turn (details in \supp~\ref{app:validation}).

\subsection{The metric checks the amount against the target}
\label{sec:val-metric}

\paragraph{Correct edits pass, wrong ones fail.}
Replaying the hidden program reproduces every target ($\Delta E \le 0.05$), and replaying it by
script passes every GUI task.
The reversed program passes 0 of 500 tasks and the closest look-alike built from tools \emph{not} in the hidden
program 1 of 500.

\paragraph{The metric is sensitive to the amount.}
Rescaled (Fig.~\ref{fig:strength}), the hidden program passes at $\times0.9$--$1.1$, on 26--48\% of the tasks at
$\times0.75$ and $\times1.25$, and never at $\times0.5$, $\times1.5$ or beyond; closeness falls smoothly on both sides.

\paragraph{VLM judges often accept the wrong amount.}
Given the target, open VLM judges misjudge 28--41\% of the known-answer candidates and even GPT-6.1 and Opus~5.5
14\% (Tab.~\ref{tab:judge}); near the boundary ($\times0.75$, $\times1.25$) all four pass 71--99\% where the metric
passes 31--36\%, and the open judges accept 11--43\% of agent outputs far from the target ($\Delta E > 6$).

\paragraph{Agreement with human judgement.}
Humans judge the 913 pairs, which cover the pass boundary from both sides and include real agent outputs
(\supp~\ref{app:val-metric}); they test whether the metric's verdicts are perceptually relevant, not whether it
predicts human preference better than a judge (Tab.~\ref{tab:judge}, right).
The metric agrees on 72.5\% of the pairs, comparable to GPT-6.1 and Opus~5.5 (72--75\%) and above the open judges
(57--68\%); away from the boundary it agrees on 82\%, and on agent outputs on 90\%.
The disagreement concentrates where the judgement is ambiguous: humans accept 80\% of the hidden programs, and at
$\times0.75$ and $\times1.25$ every scorer, metric and judges alike, agrees with them on only 48--58\%.
The best-matching threshold is 0.71 (95\% CI 0.60--0.83), which contains our 0.75; of agent outputs with
$\Delta E > 6$ humans accept 2\%, the metric none.

\begin{table}[t]
  \centering
  \small
  \caption{\textbf{Metric, VLM judges and humans on the 913 pairs of the human study.} Left: candidates with a known
  answer, share judged correctly (\%); exact: the hidden program, drop: one step left out. Right: agreement with
  human judgement (\%).
  GLM: GLM-5.3, Qwen: Qwen3.5-397B, GPT: GPT-6.1, Opus: Opus~5.5, all given the target. The metric's known-answer
  columns hold by construction (Sec.~\ref{sec:checks}); they test the judges.}
  \label{tab:judge}
  \setlength{\tabcolsep}{3.8pt}
  \begin{tabular}{@{}lccc@{\hspace{6pt}}ccc@{\hspace{6pt}}c@{\hspace{8pt}}c@{}}
    \toprule
    & \multicolumn{3}{c}{should pass} & \multicolumn{3}{c}{should fail} & & human \\
    \cmidrule(r{4pt}){2-4}\cmidrule(r{4pt}){5-7}
    & exact & $\times0.9$ & $\times1.1$ & $\times0.5$ & $\times1.5$ & drop & all & agree \\
    \midrule
    $n$    & 15  & 80  & 65  & 96 & 33 & 59  & 348 & 913 \\
    \midrule
    GLM    & 100 & 100 & 100 & 38 & 6  & 14  & 59 & 57 \\
    Qwen   & 100 & 99  & 100 & 60 & 18 & 49  & 72 & 68 \\
    GPT    & 100 & 100 & 97  & 83 & 55 & 76  & 86 & 75 \\
    Opus   & 100 & 100 & 98  & 81 & 79 & 63  & 86 & 72 \\
    Humans & 80  & 74  & 86  & 85 & 73 & 76  & 80 & -- \\
    Metric & 100 & 100 & 100 & 100 & 100 & 100 & 100 & 73 \\
    \bottomrule
  \end{tabular}
\end{table}

\subsection{The tasks are solvable, not guessable, and realistic}
\label{sec:val-tasks}

\paragraph{Solvable, but not by guessing.}
Told its distance to the target after each round (Sec.~\ref{sec:why}), GPT-6.1 passes 45--88\% of L2 tasks per
application, while a random search over the given tools with the same four renders passes only 3--6\%.

\paragraph{Generated tasks behave like real edits.}
On darktable, 160 real PlayRaw edits give the same pass rates as the generated tasks (L2 2.5\% vs.\ 2.3\%) and the
same model order by closeness (Spearman $\rho = 0.78$ at L2).

\subsection{The measurement is reliable}
\label{sec:val-models}

Our GUI harness reproduces OpenCUA-72B's published OSWorld-Verified~\cite{osworld} score (43\% vs.\ 45.0\%), and
GIMP L1 ranks the GUI agents like their published scores ($\rho = 0.90$).
L1 $\ge$ L2 $\ge$ L3 holds for every model above the floor, and rankings are stable across three seeds
(Kendall $\tau \ge 0.95$ at L1, $\ge 0.78$ by closeness at L2).

\begin{table*}[t]
  \centering
  \caption{\textbf{Main results.} Pass rate (\%) and mean closeness per level, mean over the five applications (each weighs
  the same). Program channel: 100 tasks per application and level; open models: mean of 5 seeds, closed models: one run.
  GUI channel (100 steps): 53--100 tasks per application and level; Qwen3.5 and the GUI agents: mean of 3 seeds;
  GLM-5.3, DeepSeek-V4.1, Step-3.7, Nemotron-3.5 (OSWorld's screenshot-to-pyautogui agent) and closed models: one run; episodes in which the agent left
  the application hung count as failures. Closeness: mean over episodes of
  $1-\Delta E(\text{RESULT},\text{TARGET})/\Delta E(\text{START},\text{TARGET})$, each clipped at $-1$. Pass-rate cells are shaded by the square root of their value. --: not run on this channel (the GUI agents output interface actions only). 95\% bootstrap intervals over tasks are within $\pm$3 points on programs and $\pm$4.6 on the GUI (\supp~\ref{app:ci}). Per application: Tabs.~\ref{tab:app-prog} and~\ref{tab:app-gui}.}
  \label{tab:main}
  \small\setlength{\tabcolsep}{3.5pt}\renewcommand{\arraystretch}{0.92}   %
  \begin{tabular}{@{}ll ccc@{\hspace{6pt}}ccc@{\hspace{14pt}}ccc@{\hspace{6pt}}ccc@{}}
    \toprule
    & & \multicolumn{6}{c}{Program channel} & \multicolumn{6}{c}{GUI channel} \\
    \cmidrule(r{14pt}){3-8}\cmidrule{9-14}
    & & \multicolumn{3}{c}{pass (\%)} & \multicolumn{3}{c}{closeness} & \multicolumn{3}{c}{pass (\%)} & \multicolumn{3}{c}{closeness} \\
    \cmidrule(r{6pt}){3-5}\cmidrule(r{14pt}){6-8}\cmidrule(r{6pt}){9-11}\cmidrule{12-14}
    Family & Model & L1 & L2 & L3 & L1 & L2 & L3 & L1 & L2 & L3 & L1 & L2 & L3 \\
    \midrule
    \multirow{8}{*}{Qwen3.5}     & 0.8B       & \cellcolor{CliffBlue!47}86.6 & \cellcolor{CliffBlue!6}1.4 & \cellcolor{CliffBlue!2}0.1 & 0.88 & $-$0.19 & $-$0.35 & \cellcolor{CliffBlue!3}0.4 & \cellcolor{CliffBlue!0}0.0 & \cellcolor{CliffBlue!0}0.0 & 0.01 & 0.00 & 0.00 \\
                                 & 2B         & \cellcolor{CliffBlue!48}91.9 & \cellcolor{CliffBlue!5}1.2 & \cellcolor{CliffBlue!2}0.2 & 0.93 & $-$0.15 & $-$0.41 & \cellcolor{CliffBlue!4}0.6 & \cellcolor{CliffBlue!0}0.0 & \cellcolor{CliffBlue!0}0.0 & $-$0.09 & $-$0.01 & $-$0.02 \\
                                 & 4B         & \cellcolor{CliffBlue!50}99.1 & \cellcolor{CliffBlue!11}4.6 & \cellcolor{CliffBlue!4}0.6 & 0.99 & 0.10 & $-$0.12 & \cellcolor{CliffBlue!10}4.1 & \cellcolor{CliffBlue!4}0.7 & \cellcolor{CliffBlue!0}0.0 & $-$0.02 & $-$0.01 & $-$0.17 \\
                                 & 9B         & \cellcolor{CliffBlue!50}99.9 & \cellcolor{CliffBlue!12}5.3 & \cellcolor{CliffBlue!5}0.9 & 1.00 & 0.10 & $-$0.22 & \cellcolor{CliffBlue!16}10.2 & \cellcolor{CliffBlue!4}0.5 & \cellcolor{CliffBlue!0}0.0 & 0.01 & $-$0.03 & $-$0.11 \\
                                 & 27B        & \cellcolor{CliffBlue!50}100.0 & \cellcolor{CliffBlue!14}7.7 & \cellcolor{CliffBlue!4}0.6 & 1.00 & 0.22 & $-$0.18 & \cellcolor{CliffBlue!20}16.3 & \cellcolor{CliffBlue!4}0.7 & \cellcolor{CliffBlue!2}0.2 & 0.11 & $-$0.05 & $-$0.18 \\
                                 & 35B-A3B    & \cellcolor{CliffBlue!50}100.0 & \cellcolor{CliffBlue!13}6.3 & \cellcolor{CliffBlue!5}0.9 & 1.00 & 0.18 & $-$0.16 & \cellcolor{CliffBlue!20}16.0 & \cellcolor{CliffBlue!3}0.4 & \cellcolor{CliffBlue!0}0.0 & 0.13 & $-$0.10 & $-$0.20 \\
                                 & 122B-A10B  & \cellcolor{CliffBlue!50}100.0 & \cellcolor{CliffBlue!12}6.2 & \cellcolor{CliffBlue!6}1.4 & 1.00 & 0.21 & $-$0.05 & \cellcolor{CliffBlue!20}15.3 & \cellcolor{CliffBlue!5}0.9 & \cellcolor{CliffBlue!0}0.0 & 0.12 & $-$0.04 & $-$0.13 \\
                                 & 397B-A17B  & \cellcolor{CliffBlue!50}100.0 & \cellcolor{CliffBlue!16}10.5 & \cellcolor{CliffBlue!7}2.2 & 1.00 & 0.29 & $-$0.04 & \cellcolor{CliffBlue!19}14.2 & \cellcolor{CliffBlue!5}1.0 & \cellcolor{CliffBlue!2}0.1 & 0.12 & $-$0.07 & $-$0.26 \\
    \midrule
    GLM-5.3                      & Flash      & \cellcolor{CliffBlue!50}100.0 & \cellcolor{CliffBlue!10}4.2 & \cellcolor{CliffBlue!3}0.4 & 1.00 & 0.24 & $-$0.01 & \cellcolor{CliffBlue!39}60.1 & \cellcolor{CliffBlue!11}4.7 & \cellcolor{CliffBlue!6}1.6 & 0.60 & 0.07 & $-$0.03 \\
    \midrule
    DeepSeek-V4.1                & Flash      & \cellcolor{CliffBlue!50}100.0 & \cellcolor{CliffBlue!12}5.4 & \cellcolor{CliffBlue!8}2.6 & 1.00 & 0.23 & 0.01 & \cellcolor{CliffBlue!14}8.1 & \cellcolor{CliffBlue!5}1.2 & \cellcolor{CliffBlue!3}0.4 & $-$0.02 & $-$0.09 & $-$0.07 \\
    \midrule
    Step-3.7                     & Flash      & \cellcolor{CliffBlue!50}100.0 & \cellcolor{CliffBlue!14}7.4 & \cellcolor{CliffBlue!4}0.6 & 1.00 & 0.24 & $-$0.12 & \cellcolor{CliffBlue!0}0.0 & \cellcolor{CliffBlue!0}0.0 & \cellcolor{CliffBlue!0}0.0 & $-$0.04 & $-$0.02 & $-$0.02 \\
    \midrule
    Nemotron-3.5                 & Super-VL   & \cellcolor{CliffBlue!50}100.0 & \cellcolor{CliffBlue!13}6.8 & \cellcolor{CliffBlue!4}0.6 & 1.00 & 0.22 & $-$0.04 & \cellcolor{CliffBlue!0}0.0 & \cellcolor{CliffBlue!0}0.0 & \cellcolor{CliffBlue!0}0.0 & 0.00 & $-$0.01 & $-$0.02 \\
    \midrule
    \multirow{4}{*}{UI-TARS}     & 2B         & -- & -- & -- & -- & -- & -- & \cellcolor{CliffBlue!4}0.5 & \cellcolor{CliffBlue!4}0.6 & \cellcolor{CliffBlue!0}0.0 & $-$0.03 & $-$0.03 & $-$0.10 \\
                                 & 7B         & -- & -- & -- & -- & -- & -- & \cellcolor{CliffBlue!3}0.4 & \cellcolor{CliffBlue!4}0.8 & \cellcolor{CliffBlue!0}0.0 & $-$0.05 & $-$0.03 & $-$0.08 \\
                                 & 72B        & -- & -- & -- & -- & -- & -- & \cellcolor{CliffBlue!13}6.6 & \cellcolor{CliffBlue!5}0.9 & \cellcolor{CliffBlue!0}0.0 & 0.03 & $-$0.04 & $-$0.16 \\
                                 & 1.5-7B     & -- & -- & -- & -- & -- & -- & \cellcolor{CliffBlue!16}9.8 & \cellcolor{CliffBlue!5}1.0 & \cellcolor{CliffBlue!0}0.0 & 0.05 & $-$0.05 & $-$0.24 \\
    \midrule
    \multirow{3}{*}{OpenCUA}     & 7B         & -- & -- & -- & -- & -- & -- & \cellcolor{CliffBlue!18}12.4 & \cellcolor{CliffBlue!3}0.4 & \cellcolor{CliffBlue!0}0.0 & 0.03 & $-$0.17 & $-$0.37 \\
                                 & 32B        & -- & -- & -- & -- & -- & -- & \cellcolor{CliffBlue!19}14.9 & \cellcolor{CliffBlue!3}0.3 & \cellcolor{CliffBlue!0}0.0 & 0.09 & $-$0.11 & $-$0.40 \\
                                 & 72B        & -- & -- & -- & -- & -- & -- & \cellcolor{CliffBlue!20}16.2 & \cellcolor{CliffBlue!5}0.9 & \cellcolor{CliffBlue!2}0.2 & 0.10 & $-$0.07 & $-$0.19 \\
    \midrule
    GUI-Owl-1.5                  & 32B        & -- & -- & -- & -- & -- & -- & \cellcolor{CliffBlue!18}12.3 & \cellcolor{CliffBlue!4}0.7 & \cellcolor{CliffBlue!0}0.0 & 0.08 & $-$0.10 & $-$0.39 \\
    \midrule
    EvoCUA                       & 32B        & -- & -- & -- & -- & -- & -- & \cellcolor{CliffBlue!21}17.6 & \cellcolor{CliffBlue!4}0.8 & \cellcolor{CliffBlue!0}0.0 & 0.18 & $-$0.05 & $-$0.14 \\
    \midrule
    \multirow{2}{*}{Holo4}       & 35B        & -- & -- & -- & -- & -- & -- & \cellcolor{CliffBlue!30}36.7 & \cellcolor{CliffBlue!7}1.8 & \cellcolor{CliffBlue!2}0.2 & 0.38 & $-$0.05 & $-$0.22 \\
                                 & 27B        & -- & -- & -- & -- & -- & -- & \cellcolor{CliffBlue!36}52.6 & \cellcolor{CliffBlue!8}2.3 & \cellcolor{CliffBlue!7}2.0 & 0.56 & 0.06 & $-$0.04 \\
    \midrule
    Claude                       & Opus 5.5   & \cellcolor{CliffBlue!50}100.0 & \cellcolor{CliffBlue!18}13.4 & \cellcolor{CliffBlue!13}6.8 & 1.00 & 0.45 & 0.27 & \cellcolor{CliffBlue!50}\textbf{98.8} & \cellcolor{CliffBlue!19}13.7 & \cellcolor{CliffBlue!15}8.6 & \textbf{0.98} & 0.44 & 0.31 \\
    \midrule
    GPT                          & 6.1        & \cellcolor{CliffBlue!50}100.0 & \cellcolor{CliffBlue!22}\textbf{19.0} & \cellcolor{CliffBlue!18}\textbf{13.4} & 1.00 & \textbf{0.50} & \textbf{0.38} & \cellcolor{CliffBlue!47}89.5 & \cellcolor{CliffBlue!29}\textbf{32.7} & \cellcolor{CliffBlue!21}\textbf{17.8} & 0.87 & \textbf{0.55} & \textbf{0.41} \\
    \bottomrule
  \end{tabular}
\end{table*}

\section{Experiments}
\label{sec:experiments}

\subsection{Setup}
\label{sec:setup}

\paragraph{Models.}
On the program channel we evaluate twelve open VLMs, the Qwen3.5 family~\cite{qwen35} in eight sizes (0.8B to
397B-A17B), GLM-5.3-Flash~\cite{glm5}, DeepSeek-V4.1-Flash~\cite{deepseekv41}, Step-3.7-Flash~\cite{step37} and
Nemotron-3.5-Super-VL-120B-A12B~\cite{nemotron35}, and two closed models, Claude Opus 5.5~\cite{opus55} and GPT-6.1 Sol~\cite{gpt61sol} (GPT-6.1 below), both at high
reasoning effort.
Open models are served with vLLM~\cite{vllm}; every call has an output budget of 32k tokens.
On the GUI channel we evaluate eleven GUI agents with several sizes per family, OpenCUA~\cite{opencua} (7B, 32B, 72B),
UI-TARS~\cite{uitars} (2B, 7B, 72B, 1.5-7B), GUI-Owl-1.5-32B~\cite{guiowl15}, EvoCUA-32B~\cite{evocua} and
Holo4~\cite{holo4} (27B, 35B), all twelve open VLMs and the two closed models, in the OSWorld harness~\cite{osworld}
(seeds per model in Tab.~\ref{tab:main}; \supp~\ref{app:apps}). Opus 5.5 acts through OSWorld's computer-use agent;
GPT-6.1, which has no computer-use tool on our API, and GLM-5.3, DeepSeek-V4.1, Step-3.7 and Nemotron-3.5 act through
its generic screenshot-to-pyautogui agent.

\paragraph{Protocol.}
An agent sees START (and TARGET at L2 and L3), the instruction of the level and the operation library.
It writes a program, which we render, and it sees the RESULT; it may revise and render again, up to four renders,
and decides itself when it is done.
We score the last render (Sec.~\ref{sec:metric}).
All 500 tasks are run at every level; on the 399 test tasks alone every number below changes by at most one point.
GUI agents use the interface for up to 100 steps.

\subsection{Main results}
\label{sec:main-results}

\paragraph{How we report.}
We report the pass rate, whose threshold agrees with the amount of the effect (Sec.~\ref{sec:val-metric}), and the mean
closeness, which still orders models where pass rates are near zero (Sec.~\ref{sec:val-tasks}).
Each application weighs the same, results are averaged over seeds, and GUI episodes in which the application hung count
as failures (Tab.~\ref{tab:main}; per application in Tab.~\ref{tab:app-prog} and \ref{tab:app-gui}).

\paragraph{The results behave as expected.}
The ranking follows known capability: the closed models lead on both channels, and within each open family larger
models do better, most clearly at L1 (\supp~Fig.~\ref{fig:scaling}).
Yet no model is near the ceiling: the best, GPT-6.1, passes 19\% of L2 and 13\% of L3 on programs and 33\% and 18\% on
the GUI, so the benchmark leaves ample room to separate future agents.
The GUI is harder for open models and GUI agents (L1 0--60\%), and Inkscape is the easiest application (\supp~\ref{app:results}).

\subsection{Agents operate, but do not reach the effect}
\label{sec:cliff}

\paragraph{The cliff.}
Told which tools to use and with which values (L1), every program model from 4B up passes 99--100\%: writing a valid
program in an unfamiliar library is not the problem.
Told only the tools (L2), the open models pass 1.2--10.5\% and the closed models 13--19\%; told only the goal (L3), at
most 2.6\% and 7--13\%. At L3 eleven of the twelve open models have negative mean closeness: their edits move the image
\emph{away} from the target.

\paragraph{Not the interface.}
The closed models operate the GUI almost as well as they write programs (L1 99\% and 90\%), yet fall off the same cliff
(L2 14\% and 33\%, L3 9\% and 18\%). So does the best open model on the GUI, GLM-5.3 (L1 60\%, L2 4.7\%,
L3 1.6\%); DeepSeek-V4.1 passes 8\% at L1, Step-3.7 and Nemotron-3.5 none. For every model that operates the GUI, the gap is the effect, not operation.

\paragraph{Size raises operation, not judgement.}
Within OpenCUA, UI-TARS and Qwen3.5, GUI L1 rises with size, while L2 stays at most 1\% and L3 at most 0.2\%
(\supp~Fig.~\ref{fig:scaling}). On programs, Qwen3.5 L2 rises only from 1.4\% to 10.5\% over a 500-fold range of sizes, while L1 saturates
from 4B.

\begin{table*}[t]
  \centering
  \caption{\textbf{What changes L2 pass rates.} Pass rate (\%) of the default run and of the run with the addition, on the
  same episodes, and the change $\Delta$ (percentage points); \textbf{bold}: significant ($p<0.05$, McNemar),
  \textcolor{gray}{gray}: not significant. Open: the twelve open models (first two blocks: 30-task subsets per application); Opus 5.5 / GPT-6.1: the same subsets in the first two blocks, all 500
  tasks in the last. Image of the difference: a rendered map of TARGET $-$ START per Lab channel instead of its statistics. --: not run.
  Closeness changes and $p$-values: Tab.~\ref{tab:interventions-full}.}
  \label{tab:interventions}
  \setlength{\tabcolsep}{4pt}
  \resizebox{\textwidth}{!}{%
  \begin{tabular}{@{}l ccc@{\hspace{12pt}}ccc@{\hspace{12pt}}ccc@{}}
    \toprule
    & \multicolumn{3}{c}{Open} & \multicolumn{3}{c}{Opus 5.5} & \multicolumn{3}{c}{GPT-6.1} \\
    \cmidrule(r{12pt}){2-4}\cmidrule(r{12pt}){5-7}\cmidrule{8-10}
    Added to the input & Default & Added & $\Delta$ & Default & Added & $\Delta$ & Default & Added & $\Delta$ \\
    \midrule
    \multicolumn{10}{@{}l}{\emph{Knowledge about the tools: little help}} \\
    Full application manual                    & 5.7 & 4.7 & \textcolor{gray}{$-$1.0} & 12.7 & 14.0 & \textcolor{gray}{$+$1.3} & 17.3 & 16.7 & \textcolor{gray}{$-$0.6} \\
    Reference card per tool                    & 5.7 & 6.3 & \textcolor{gray}{$+$0.6} & 12.7 & 16.0 & \textcolor{gray}{$+$3.3} & 17.3 & 24.0 & \textbf{$+$6.7} \\
    Cards + each tool's effect on an image     & 5.7 & 6.8 & \textcolor{gray}{$+$1.1} & 12.7 & 13.3 & \textcolor{gray}{$+$0.6} & 17.3 & 25.3 & \textbf{$+$8.0} \\
    Each setting's effect measured on START    & 5.7 & 4.3 & \textbf{$-$1.4} & 12.7 & 14.0 & \textcolor{gray}{$+$1.3} & 17.3 & 25.3 & \textbf{$+$8.0} \\
    \midrule
    \multicolumn{10}{@{}l}{\emph{The measured difference: large gains}} \\
    Statistics of TARGET $-$ START             & 5.7 & 14.9 & \textbf{$+$9.2} & 12.7 & 34.0 & \textbf{$+$21.3} & 17.3 & 47.3 & \textbf{$+$30.0} \\
    \quad the same, as an image of the difference & 5.7 & 6.9 & \textcolor{gray}{$+$1.2} & -- & -- & -- & -- & -- & -- \\
    \quad the same, GUI channel                & -- & -- & -- & 10.4 & 30.4 & \textbf{$+$20.0} & 28.0 & 39.2 & \textbf{$+$11.2} \\
    \midrule
    \multicolumn{10}{@{}l}{\emph{Looking again: helps strong models}} \\
    Four forced rounds, each RESULT shown      & 5.6 & 8.2 & \textbf{$+$2.6} & 13.4 & 29.8 & \textbf{$+$16.4} & 19.0 & 50.4 & \textbf{$+$31.4} \\
    \quad + $\Delta E$ to TARGET each round    & 5.6 & 13.2 & \textbf{$+$7.6} & 13.4 & 38.0 & \textbf{$+$24.6} & 19.0 & 62.2 & \textbf{$+$43.2} \\
    \bottomrule
  \end{tabular}}
\end{table*}

\paragraph{Not the threshold.}
With the loosest rule ($\Delta E \le 4$ alone; it also passes 34\% of the $\times0.5$ and $\times1.5$ programs), open models still pass at most 32\% (L2) and 11\% (L3) (\supp~\ref{app:thresholds}).

\subsection{Why agents fail}
\label{sec:why}

At L2 the tools are given; what remains is how much of each to apply, to be read from START and TARGET.
We vary what the agent is given (Tab.~\ref{tab:interventions}).

\paragraph{Not missing tool knowledge.}
Knowledge about the tools helps little: the application manual changes no model's pass rate, and reference cards,
example images or the measured effect of each setting help only GPT-6.1, modestly (17\% $\to$ 24--25\%), while they
lower the open models' closeness.

\paragraph{Seeing the difference is the main bottleneck.}
Given the statistics of TARGET $-$ START, the closed models more than double their pass rate (Opus 5.5 13\% $\to$ 34\%,
GPT-6.1 17\% $\to$ 47\%), also on the GUI (Opus 5.5 10\% $\to$ 30\%, GPT-6.1 28\% $\to$ 39\%), and so do the open models
(5.7\% $\to$ 14.9\%; significant for 9 of the 12, Qwen3.5-27B 7\% $\to$ 29\%; not for the two smallest Qwen3.5 sizes and Step-3.7).
Shown the same difference as an image instead (a map of TARGET $-$ START per Lab channel), the open models gain nothing
(5.7\% $\to$ 6.9\%, n.s.). The models can map a measured difference onto the tools, if imperfectly, but the open models fail to read it from
a single look at the images, even from a map of the difference.

\paragraph{Agents do not check their result.}
Allowed four renders, Opus 5.5 and GPT-6.1 stop after the first in 99.8\% and 100\% of the L2 episodes (GLM-5.3 98\%,
Qwen3.5-397B 84\%), accepting results a median $\Delta E$ of 5--7.5 from the target.

\paragraph{Forced comparison helps strong models.}
Made to use all four rounds, the closed models more than double their L2 pass rate (Opus 5.5 13\% $\to$ 30\%, GPT-6.1
19\% $\to$ 50\%), while the open models gain little (5.6\% $\to$ 8.2\%). The error falls round by round
(Fig.~\ref{fig:rounds}): the models use each RESULT to correct the next program. Adding the distance after each round helps
further (open 13\%, Opus 5.5 38\%, GPT-6.1 62\%); maximum reasoning effort helps about as much as forced rounds (Opus 5.5
13\% $\to$ 31\% on 30-task subsets). At L3 every model stays far from solved, even with the distance (at most 37\%).

\paragraph{Operating and judging are separate skills.}
On the GUI, Opus 5.5 operates better and GPT-6.1 judges better (Sec.~\ref{sec:cliff}); GPT-6.1's L1 failures are
operational (\supp~\ref{app:results}).

\section{Related Work}
\label{sec:related}

\paragraph{Benchmarks for computer-use agents.}
OSWorld~\cite{osworld} established executable tasks in real desktop applications with a script that checks the final
state, and its GIMP tasks are among the few on creative software.
Later benchmarks target professional software more directly: PSBench~\cite{psbench} for photo editing,
OSExpert~\cite{osexpert} and CUA-World~\cite{cuaworld} for expert workflows across applications,
ProSoftArena~\cite{prosoftarena} for scientific and industrial software, CutVerse~\cite{cutverse} for media
post-production, and BlenderGym~\cite{blendergym} for 3D scene edits towards a goal image.
They score a scripted check of the application state, which needs explicit tasks, or a VLM judge, which cannot tell an effect's amount (Sec.~\ref{sec:motivation}).
We keep their executable setting and interface and add tasks whose answer is an effect a reference metric can verify.
Unlike them, every task has a known target and is posed at three levels, so a failure can be traced to operation or
to judgement.
Retouching agents~\cite{jarvisart,monetgpt} and image-to-code models~\cite{starvector} choose operations and their
values; our tasks measure exactly how well those values are chosen.

\paragraph{GUI agents.}
Open GUI agents such as OpenCUA~\cite{opencua} and UI-TARS~\cite{uitars} are trained on large collections of
interaction traces and evaluated on operation: finding and using the right control~\cite{seeclick,cogagent,webarena,mind2web}.
Our L1 level measures the same skill and agrees with their published ranking (Sec.~\ref{sec:val-models}); L2 and L3
show that it does not carry over to judging the effect.
Self-refinement improves LLMs on text tasks~\cite{selfrefine,reflexion,selfcorrect}; editing agents rarely re-check their
render, and forced comparison helps only strong models (Sec.~\ref{sec:why}).

\paragraph{Evaluating image edits.}
Instruction-based image editing~\cite{instructpix2pix,emuedit,step1xedit} is usually evaluated by VLM judges~\cite{llmjudge,viescore} or by CLIP-style similarity to the instruction~\cite{magicbrush,imagenhub}.
These measure whether the right kind of change happened, not whether its amount matches; reference-based colour
metrics such as CIEDE2000~\cite{ciede2000} do measure the amount but need a reference.
Our generated tasks provide one by construction, because the target is rendered from a known program.

\paragraph{Fine-grained visual comparison in VLMs.}
VLMs struggle with fine-grained and low-level perception and with comparing two similar images~\cite{mmvp,blink,vlmblind,qbench,compbench}.
These benchmarks ask questions about fixed images; in \bench the model must act on the difference until it reaches the target amount.
Our results locate this weakness in a practical setting: given the measured difference, models apply it, but cannot read it from a single look at the images (Sec.~\ref{sec:why}).

\section{Conclusion}
\label{sec:conclusion}

Requests for professional editing describe an effect, not the steps to reach it.
\bench tests whether an agent reaches a described effect in five applications and two channels, with a known
target for every task and a metric that, unlike VLM judges, checks the amount of the effect.
Agents operate the tools but miss the amount: they apply a measured difference yet cannot read it from a single look at the images,
and they do not check their result. Effect judgement, not operation, is what agents now need to learn.

\paragraph{Limitations.}
\bench scores an edit by its rendered appearance, so it covers software whose result is visual; non-visual tools,
such as audio, would need another reference metric (\supp~\ref{app:limitations}).

\section*{LLM Usage Statement}

\textbf{As part of the method.}
Two generator LLMs (Qwen3.5-397B and GLM-5.3) propose the editing intent and the edit program of every task and judge
whether a target is plausible (Sec.~\ref{sec:benchmark}); every task is then checked by rendering and by fixed rules,
not by the LLM's opinion.
Two LLMs label forum edit requests as explicit or vague in the motivation study (Sec.~\ref{sec:motivation}).
VLM judges appear only as baselines that we compare against our metric, never as the evaluator of a reported score. \textbf{As subjects of study.}
All agents evaluated in the paper are LLM- or VLM-based (open models served locally; Claude Opus~5.5 and GPT-6.1
through a commercial API). \textbf{As research and writing assistants.}
An LLM coding assistant was used to write and debug experiment code (environment adapters, run pipelines, scoring and
plotting scripts) and to operate the compute runs; all code paths that produce reported numbers were reviewed by the
authors, and all numbers are computed by scripts from the stored episodes. LLMs were also used to edit the text for
clarity and length. The authors designed the study, checked every claim and number against the data, and take full
responsibility for the content.

{
    \small
    \bibliographystyle{ieeenat_fullname}
    \bibliography{main}
}

\clearpage
\appendix
\setcounter{figure}{0}\setcounter{table}{0}
\renewcommand{\thefigure}{S\arabic{figure}}\renewcommand{\theHfigure}{S\arabic{figure}}
\renewcommand{\thetable}{S\arabic{table}}\renewcommand{\theHtable}{S\arabic{table}}
\raggedbottom
\renewcommand{\topfraction}{0.9}\renewcommand{\dbltopfraction}{0.9}
\renewcommand{\bottomfraction}{0.6}\renewcommand{\textfraction}{0.08}
\renewcommand{\floatpagefraction}{0.7}\renewcommand{\dblfloatpagefraction}{0.7}
\setcounter{topnumber}{4}\setcounter{dbltopnumber}{4}\setcounter{bottomnumber}{2}\setcounter{totalnumber}{6}
\makeatletter
\setlength{\@fptop}{0pt}\setlength{\@fpbot}{0pt plus 1fil}
\setlength{\@dblfptop}{0pt}\setlength{\@dblfpbot}{0pt plus 1fil}
\makeatother

\twocolumn[{\centering\Large\bfseries Appendix\par\vspace{1em}}]
\noindent Sections A--E, Figs.~S\# and Tabs.~S\# belong to this appendix; numbered sections (1--7), figures and
tables without a prefix to the main paper.

\smallskip
\noindent\textbf{Contents.}
\newcommand{\tocline}[2]{\item[\ref{#1}] #2 \dotfill\ \pageref{#1}}
\begin{itemize}[leftmargin=*,topsep=1pt,itemsep=0pt]
  \tocline{app:motivation}{\textbf{Motivation experiments: setups and full results}}
  \begin{itemize}[leftmargin=2.2em,topsep=0pt,itemsep=0pt]
    \tocline{app:r1}{Real requests}
    \tocline{app:e1}{Task phrasing}
    \tocline{app:e1b}{Agent success by task class}
    \tocline{app:e2}{Evaluator audit}
    \tocline{app:e3}{Judge stress test}
    \tocline{app:e3x}{Judge stress test on darktable}
    \tocline{app:c1}{Editing without the software}
  \end{itemize}
  \tocline{app:benchmark}{\textbf{Benchmark details}}
  \begin{itemize}[leftmargin=2.2em,topsep=0pt,itemsep=0pt]
    \tocline{app:stats}{Dataset statistics}
    \tocline{app:apps}{Applications and environment}
    \tocline{app:generation}{Generation details}
    \tocline{app:levels}{Instructions}
    \tocline{app:ops}{Operation libraries}
    \tocline{app:drops}{Drop logs}
  \end{itemize}
  \tocline{app:validation}{\textbf{Validation details}}
  \begin{itemize}[leftmargin=2.2em,topsep=0pt,itemsep=0pt]
    \tocline{app:val-metric}{Metric}
    \tocline{app:val-tasks}{Tasks}
    \tocline{app:val-models}{Ranking}
  \end{itemize}
  \tocline{app:results}{\textbf{Results per application}}
  \begin{itemize}[leftmargin=2.2em,topsep=0pt,itemsep=0pt]
    \tocline{app:per-app}{Per application}
    \tocline{app:rounds}{Error after each round}
    \tocline{app:thresholds}{Other pass rules}
    \tocline{app:interventions}{Interventions}
  \end{itemize}
  \tocline{app:limitations}{\textbf{Limitations}}
\end{itemize}

\section{Motivation experiments: setups and full results}
\label{app:motivation}

This appendix gives the setup and the full results of every experiment in Sec.~\ref{sec:motivation}
(Fig.~\ref{fig:motivation} is drawn from the same results).

\subsection{Real requests}
\label{app:r1}
We collected 960 public posts, 240 from each of four communities:
r/PhotoshopRequest (raster edits), r/editmyraw (RAW grading), and the GIMP and darktable forums
(Reddit and discuss.pixls.us).
Posts span Oct.\ 2024--Sep.\ 2026, drawn evenly per month (Reddit, through the Arctic Shift archive)
or from random pages of the category listing (pixls); we use the text only.
Every edit clause is labelled by two LLMs (Qwen3.5-397B, GLM-5.3) with a fixed rule:
\emph{explicit} if a real command is named and its values are exact or not needed;
\emph{missing values} if the command is named but not its amount;
\emph{missing tool} if no command is named.
Posts with no edit request (bug reports, installation questions, showcases) are dropped.
Tab.~\ref{tab:app-r1} reports shares over the requests on which both labellers agree.
The software forums are mostly bug and how-to posts, so they contain few edit requests and the labellers agree less;
the claim rests on the two request communities.
Counting only agreed labels keeps the shares tied to the fixed rule rather than to one model's reading of a post.

\begin{table*}[!htbp]
  \centering
  \small
  \caption{\textbf{Real requests.} Shares (\%) over the edit requests on which both labellers agree.}
  \label{tab:app-r1}
  \setlength{\tabcolsep}{3.5pt}
  \begin{tabular}{@{}lccccc@{}}
    \toprule
    Source & Posts & Edit req. & Agree & Missing tool & Explicit \\
    \midrule
    r/PhotoshopRequest & 240 & 211 & 94 & 99  & 0 \\
    r/editmyraw        & 240 & 38  & 77 & 100 & 0 \\
    GIMP forums        & 240 & 30  & 72 & 97  & 3 \\
    darktable forums   & 240 & 9   & 65 & 89  & 0 \\
    \bottomrule
  \end{tabular}
\end{table*}

\subsection{Task phrasing}
\label{app:e1}
We applied the same rule to the editing tasks of PSBench (600 tasks), OSWorld (26 GIMP tasks), OSExpert (20) and
CUA-World (83).
For PSBench three LLM voters (Qwen3.5-397B, GLM-5.3, DeepSeek-V4.1-Flash) label every clause and the majority class
is used; they are unanimous on 82\% of the tasks and split three ways on 3 of 600.
For the other benchmarks we add a \emph{UI-operation} class (non-visual actions such as changing a setting) and a
\emph{no-edit} class; two voters label them, and we report shares over the edit tasks for each voter.

\begin{table}[!htbp]
  \centering
  \small
  \caption{\textbf{Task phrasing.} Share (\%) of edit tasks per class. PSBench: majority of three voters.
  Others: Qwen3.5 / GLM-5.3.}
  \label{tab:app-e1}
  \setlength{\tabcolsep}{3.5pt}
  \begin{tabular}{@{}lccc@{}}
    \toprule
    Benchmark & Explicit & Missing values & Missing tool \\
    \midrule
    PSBench easy (200)   & 57      & 22 & 21 \\
    PSBench medium (203) & 12      & 21 & 67 \\
    PSBench hard (197)   & 0       & 2  & 98 \\
    OSWorld GIMP         & 6 / 0   & 31 / 31 & 62 / 69 \\
    OSExpert             & 33 / 29 & 28 / 35 & 39 / 35 \\
    CUA-World            & 16 / 18 & 9 / 10  & 76 / 72 \\
    \bottomrule
  \end{tabular}
\end{table}

\subsection{Agent success by task class}
\label{app:e1b}
We joined the task-phrasing classes of the 26 OSWorld GIMP tasks with the published OSWorld-Verified results
(94 agents, 106--107 runs per task) and averaged the per-task success within each class,
with a task-level bootstrap 95\% CI.
``Feasible'' drops the 9 tasks whose evaluator only checks that the agent refuses an infeasible request.
Two to three of the vague tasks are scored by direction-only checks, so their success is an upper bound.

\begin{table}[!htbp]
  \centering
  \small
  \caption{\textbf{Agent success by task class.} Mean success (\%) over published agents, OSWorld GIMP, feasible tasks; [95\% CI] (n tasks).}
  \label{tab:app-e1b}
  \setlength{\tabcolsep}{3pt}
  \begin{tabular}{@{}lccc@{}}
    \toprule
    Labels & UI operation & Missing values & Missing tool \\
    \midrule
    adjudicated & 85 [79, 90] (5) & 72 (2)          & 39 [20, 72] (3) \\
    Qwen3.5     & 85 [79, 90] (5) & 63 [39, 77] (4) & 47 [28, 67] (5) \\
    GLM-5.3     & 78 [62, 89] (6) & 66 [57, 72] (4) & 49 [30, 67] (6) \\
    \bottomrule
  \end{tabular}
\end{table}

\subsection{Evaluator audit}
\label{app:e2}
We read the released evaluator code of every benchmark discussed in Sec.~\ref{sec:related}, and the paper where no code is
released.
PSBench scores results with a VLM judge that is never shown the reference result;
OSWorld's ``increase/decrease'' checks test only the direction of a change;
16 of the 20 OSExpert GIMP evaluators are not released.
Only OSWorld's five reference-metric GIMP tasks and ProSoftArena's Photoshop MSE compare against a target.

\subsection{Judge stress test}
\label{app:e3}
For 60 PSBench tasks (explicit and vague) we built, from PSBench's reference result (END), controlled candidates:
the edit at half strength ($\times 0.5$) and triple strength ($\times 3$), the opposite edit,
a naive global edit, and the unedited start image (identity).
Each candidate is scored with the PSBench judge prompt, verbatim, by five open VLM judges
(Qwen3.5, GLM-5.3, DeepSeek-V4.1, Step-3.7, Nemotron-3.5), three repeats each.
In a control run two judges (Qwen3.5, GLM-5.3) are also shown END and told to judge strength against it;
near-correct candidates ($\times 0.9$, $\times 1.1$) are added there so that over-strictness is visible.
The reference metric is CIEDE2000 $\le 3$ (mean, images at $\le 1024$ px).
Fig.~\ref{fig:motivation}c pools judges, repeats and task types.

\begin{table*}[!tp]
  \centering
  \small
  \caption{\textbf{Judge stress test.} Pass rate (\%). No reference: five judges, explicit / vague tasks.
  With reference: two judges, pooled.}
  \label{tab:app-e3}
  \setlength{\tabcolsep}{3pt}
  \begin{tabular}{@{}lccccccc@{}}
    \toprule
    Scorer & END & $\times$0.9 / 1.1 & $\times 0.5$ & $\times 3$ & opp. & naive & ident. \\
    \midrule
    Judge, no ref.   & 82 / 86 & --      & 45 / 64 & 52 / 63 & 25 / 26 & 42 / 32 & 3 / 8 \\
    Judge, with ref. & 91      & 86 / 88 & 22      & 20      & 5       & 22      & 1 \\
    CIEDE2000 $\le 3$ & 100    & 87 / 90 & 18      & 5       & 2       & 22      & 5 \\
    \bottomrule
  \end{tabular}
\end{table*}
Every judge passes 33--73\% of the wrong-magnitude results.
With the reference the judge still fails 9\% of exact results and passes 20--22\% of wrong magnitudes,
and this depends strongly on the judge (GLM-5.3: 34--36\%; Qwen3.5: 6--8\%).

\subsection{Judge stress test on darktable}
\label{app:e3x}
The 56 darktable tasks of our pilot are posed as PSBench task folders, each twice:
explicit (our L1 text, modules and exact values) and vague (a client-style request written by Qwen3.5 from START
and TARGET: the look only, no tools or numbers).
The rubric follows PSBench's rule (one criterion per sentence).
Candidates are real darktable renders of the hidden program scaled $\times 1$ (end), $\times 0.5$ (half),
$\times 3$ (over), $\times -1$ (opposite), plus identity; clipped renders are dropped,
which leaves too few ``over'' and ``opposite'' candidates to report.
Two judges (Qwen3.5, GLM-5.3), three repeats each. Our metric is defined in Sec.~\ref{sec:metric}.

\begin{table*}[!htbp]
  \centering
  \small
  \caption{\textbf{Judge stress test on darktable.} Pass rate (\%) on darktable renders, [95\% CI].}
  \label{tab:app-e3x}
  \setlength{\tabcolsep}{3.5pt}
  \begin{tabular}{@{}llcccc@{}}
    \toprule
    Judge & Instruction & identity & $\times 0.5$ & END & our metric ($\times 0.5$ / END) \\
    \midrule
    GLM-5.3 & explicit & 16 & 98 [96, 100] & 98 & 0 / 100 \\
    GLM-5.3 & vague    & 10 & 82 [74, 90]  & 92 & 0 / 100 \\
    Qwen3.5 & explicit & 0  & 59 [48, 70]  & 88 & 0 / 100 \\
    Qwen3.5 & vague    & 1  & 65 [53, 76]  & 92 & 0 / 100 \\
    \bottomrule
  \end{tabular}
\end{table*}

\subsection{Editing without the software}
\label{app:c1}
Qwen-Image-Edit-2511 (diffusers, model-card settings: 40 steps, true CFG scale 4.0, seed 0) edits each of the 600
PSBench start images from the instruction alone; the output is resized to the start image's size.
Outputs are scored with the PSBench judge prompt by Qwen3.5 and GLM-5.3, three repeats each.
The unedited start image and the reference result are the lower and upper controls.
By task-phrasing class, the image model passes 80\% of the missing-tool tasks with both judges, against 44--61\% of the explicit
tasks, whose exact result the generator does not reproduce.

\begin{table}[!htbp]
  \centering
  \small
  \caption{\textbf{Editing without the software.} PSBench judge pass rate (\%), Qwen3.5 / GLM-5.3 as judge.}
  \label{tab:app-c1}
  \setlength{\tabcolsep}{4pt}
  \begin{tabular}{@{}lccc@{}}
    \toprule
    Difficulty & Start image & Image model & Reference result \\
    \midrule
    Easy (200)   & 4 / 17 & 64 / 76 & 92 / 96 \\
    Medium (203) & 5 / 11 & 47 / 62 & 89 / 83 \\
    Hard (197)   & 5 / 14 & 96 / 86 & 97 / 93 \\
    \bottomrule
  \end{tabular}
\end{table}

\section{Benchmark details}
\label{app:benchmark}

\subsection{Dataset statistics}
\label{app:stats}

\begin{table}[!htbp]
  \centering
  \small
  \caption{\textbf{Dataset statistics.} Tasks (dev / test), operation-library size, median number of parameter
  values per task, median CIEDE2000 between START and TARGET, share of generated candidates that passed every check,
  and tasks in the GUI channel (those whose program can be replayed through the interface).
  Held-out means that test tasks were not used during development; their answers are released with the benchmark.}
  \label{tab:bench-stats}
  \setlength{\tabcolsep}{3.2pt}
  \begin{tabular}{@{}lcccccc@{}}
    \toprule
    App & Dev / test & Ops & Val. & $\Delta E$ & Kept & GUI \\
    \midrule
    darktable      & 20 / 80 & 9  & 6  & 9.0  & 11\% & 53 \\
    RawTherapee    & 20 / 80 & 11 & 6  & 7.7  & 21\% & 100 \\
    GIMP           & 20 / 80 & 11 & 5  & 8.6  & 5\%  & 100 \\
    Inkscape       & 21 / 79 & 11 & 3  & 8.1  & 5\%  & 73 \\
    Blender        & 20 / 80 & 7  & 8  & 13.4 & 16\% & 100 \\
    \midrule
    PlayRaw (real) & 160 tasks & 10 & 19 & 12.0 & 51\% & -- \\
    \bottomrule
  \end{tabular}
\end{table}

\subsection{Applications and environment}
\label{app:apps}
Versions are pinned: darktable~5.6.2, RawTherapee~5.13, GIMP~3.2.6, Inkscape~1.4.4 and Blender~4.5.
The environment follows the OSWorld interface (\texttt{reset}, \texttt{step}, \texttt{evaluate}).
Each adapter loads the start state, reads the edit state back from the project file (a darktable XMP history, a
RawTherapee processing profile, a GIMP filter stack, SVG filter attributes, Blender node parameters), renders it with
the application's headless renderer and lists the operation library.
In the program channel the agent may render up to four times; in the GUI channel it has up to 100 steps in a virtual
desktop, and the evaluator reads the project the application saved.

\paragraph{GUI harness.}
Agents, prompts and reply parsing are OSWorld's own; each step's action (pyautogui code, or a computer-use tool call
for Claude) is executed on the episode's virtual X display. Unlike OSWorld, which runs every episode in its own VM, we run
many episodes side by side on one host, so the executor isolates them: imports in the agent's code are bound to the
episode's display (\texttt{import pyautogui as p}, \texttt{subprocess} and clipboard calls included), commands that kill
processes are refused, and file searches from the root are confined to the episode's home directory. The harness
reproduces OpenCUA-72B's published OSWorld-Verified score (App.~\ref{app:val-models}).

\subsection{Generation details}
\label{app:generation}
\paragraph{Assets.}
Photographs from PSBench~\cite{psbench} for darktable, RawTherapee and GIMP, public-domain Openclipart drawings for
Inkscape, Poly Haven scenes for Blender; an asset is used at most once per application, and the 500 tasks use 417
distinct assets (the photo applications share some photographs).
\paragraph{Generator.}
The generator LLM is given the start asset, the operation library and the manual's workflow pages, and writes an
intent (\eg ``make the portrait pop with a warm, sunlit look''), a program and an interval for every value.
Qwen3.5-397B and GLM-5.3 alternate and produce 47--53\% of the tasks each.
\paragraph{Check thresholds.}
Exact: two renders and the replay agree within $\Delta E \le 0.05$.
Visible: START--TARGET $\Delta E$ between 6 and 30 (up to 45 for five steps; a lower floor for sharpening), and the
edit does not blow out or crush a growing share of pixels.
Needed and Sensitive use the pass rule of Sec.~\ref{sec:metric}; Sensitive skips a strength that runs into a
parameter bound.
Not trivial: a shift of START to TARGET's mean colour and a Reinhard colour transfer must fail.
\paragraph{Revision and freezing.}
When a candidate fails on effect size or on a step that adds too little, the measured numbers are returned to the
generator, which revises its intervals up to two times; otherwise the candidate is dropped and the slot is retried
with a new asset.
Doubtful targets are reviewed visually before the sets are frozen.
\paragraph{Coverage.}
Program lengths follow an equal quota of one to five steps; five-step tasks are rarer in GIMP and Inkscape (10 and 8).
All but two library operations appear in at least seven tasks (GIMP's Gaussian blur once, Blender's gamma never).
Only 5\% of the GIMP and Inkscape candidates passed every check, against 11--21\% for the other applications.
\paragraph{Real tasks.}
The PlayRaw edits (discuss.pixls.us) are replayed in our action space and kept only if they reproduce the user's
result and pass the same checks. They are longer (mostly 2--7 steps, a median of 19 parameter values) and further from
START; since their assets are public they may appear in pre-training data, so we report them separately.

\subsection{Instructions}
\label{app:levels}
All instructions are generated from the hidden program by a template; none is written by hand or by an LLM, so the
wording itself is never vague.
Every level starts with the same header, which lists the operations the agent may use with their fields and ranges
(the task's operations at L1 and L2, the whole library at L3), followed by the level-specific text shown in
Fig.~\ref{fig:task-example}.

\subsection{Operation libraries}
\label{app:ops}
Tab.~\ref{tab:ops} lists each application's library. Names are the GUI labels, so the same library serves the program
and the GUI channel.

\begin{table*}[!t]
  \centering
  \footnotesize
  \caption{\textbf{Operation libraries.} Operations an agent may use at L3 (GUI labels), with the fields it can set in
  parentheses; ranges are in the released task files.}
  \label{tab:ops}
  \setlength{\tabcolsep}{4pt}
  \renewcommand{\arraystretch}{1.15}
  \begin{tabular}{@{}l c p{0.82\textwidth}@{}}
    \toprule
    App & Ops & Operations (fields) \\
    \midrule
    darktable & 9 & exposure (exposure); contrast brightness saturation (contrast, brightness, saturation); levels (levels); shadows and highlights (shadows, highlights); color balance rgb (shadows\_H, shadows\_C, highlights\_H, highlights\_C, global\_H, global\_C, saturation\_global, vibrance, contrast); local contrast (detail); sharpen (amount, radius); vignetting (brightness, scale); haze removal (strength). \\
    \addlinespace[2pt]
    RawTherapee & 11 & Exposure (compensation, contrast, saturation); L*a*b* Adjustments (brightness, chromaticity); Shadows/Highlights (highlights, shadows); Vibrance (pastels, saturated); Local Contrast (amount, radius); Sharpening (amount, radius); White Balance (temperature, green); Haze Removal (strength); Vignette Filter (strength, feather, roundness); Soft Light (strength); Color Toning (L*a*b* color correction grid) (shadows\_a, shadows\_b, highlights\_a, highlights\_b). \\
    \addlinespace[2pt]
    GIMP & 11 & Colors $>$ Brightness-Contrast (brightness, contrast); Colors $>$ Exposure (exposure, black\_level); Colors $>$ Color Temperature (intended\_temperature); Colors $>$ Hue-Saturation (hue, saturation, lightness); Colors $>$ Color Balance (cyan\_red, magenta\_green, yellow\_blue); Colors $>$ Levels (low\_input, high\_input, gamma); Colors $>$ Shadows-Highlights (shadows, highlights); Filters $>$ Enhance $>$ Sharpen (Unsharp Mask) (radius, amount); Filters $>$ Blur $>$ Gaussian Blur (size); Filters $>$ Light and Shadow $>$ Vignette (radius, softness); Filters $>$ Artistic $>$ Softglow (glow\_radius, brightness). \\
    \addlinespace[2pt]
    Inkscape & 11 & Blur (Fill and Stroke) (radius); Opacity (Fill and Stroke) (opacity); Hue Rotate (Filter Editor: Color Matrix) (angle); Saturate (Filter Editor: Color Matrix) (amount); Brightness (Filter Editor: Component Transfer, linear) (offset); Contrast (Filter Editor: Component Transfer, linear) (amount); Gamma (Filter Editor: Component Transfer, gamma) (exponent); Colorize (Filters $>$ Color $>$ Colorize) (hue, strength); Posterize (Filters $>$ Color $>$ Posterize) (levels); Drop Shadow (Filters $>$ Shadows and Glows) (offset, blur, opacity); Roughen (Filters $>$ Distort) (scale). \\
    \addlinespace[2pt]
    Blender & 7 & Color Management $>$ Exposure (ev); Color Management $>$ Gamma (gamma); Color Management $>$ White Balance (temperature, tint); World $>$ Background (HDRI) (strength, rotation); Add $>$ Light $>$ Sun (strength, elevation, azimuth); Ground plane $>$ Material (Principled BSDF Base Color / Roughness) (hue, saturation, value, roughness); Add $>$ UV Sphere + Principled Volume (fog volume around camera and scene) (density). \\
    \bottomrule
  \end{tabular}
\end{table*}

The PlayRaw tasks use the darktable modules found in the users' edits that our adapter can replay
(10 modules, including AgX tone mapping, the tone equalizer and the colour equalizer).

\subsection{Drop logs}
\label{app:drops}
Tab.~\ref{tab:drops} lists, for every application, the generated candidates and the first check each rejected
candidate failed (checks run in this order; revisions inside a candidate are not counted).
All attempts and reasons are released with the benchmark.

\begin{table*}[!tp]
  \centering
  \small
  \caption{\textbf{Generated candidates by first failed check.}
  Visible: START--TARGET $\Delta E$ out of range or clipping; Needed: removing a step does not fail;
  Sensitive: $\times0.9$/$\times1.1$ fail or $\times0.5$/$\times1.5$ pass;
  Not trivial: a mean-colour shift or Reinhard transfer passes; Other: render errors, unusable assets, operation-coverage cap.}
  \label{tab:drops}
  \setlength{\tabcolsep}{2.5pt}
  \begin{tabular}{@{}lrrrrrrrr@{}}
    \toprule
    App & Cand. & Visible & Needed & Sensitive & Not triv. & Plausible & Other & Kept \\
    \midrule
    darktable   & 1{,}061 & 306 & 389 & 47  & 70  & 60 & 69  & 120 \\
    RawTherapee & 595     & 174 & 150 & 26  & 62  & 33 & 27  & 123 \\
    GIMP        & 2{,}363 & 680 & 844 & 243 & 336 & 63 & 80  & 117 \\
    Inkscape    & 2{,}166 & 363 & 1{,}152 & 169 & 259 & 27 & 82 & 114 \\
    Blender     & 908     & 58  & 242 & 218 & 111 & 24 & 110 & 145 \\
    \midrule
    Total       & 7{,}093 & 1{,}581 & 2{,}777 & 703 & 838 & 207 & 368 & 619 \\
    \bottomrule
  \end{tabular}
\end{table*}

Kept candidates exceed 100 per application because some were removed afterwards, mainly in the visual review,
and their slots were filled again.

\section{Validation details}
\label{app:validation}

\subsection{Metric}
\label{app:val-metric}

\paragraph{Replay.}
For every task the hidden program is written into the application's edit state and re-rendered headless; the render
matches TARGET with $\Delta E \le 0.05$ on all 500 tasks.
For the GUI channel the program is replayed through the interface by a script on all 426 GUI tasks; every replay
passes, with a median $\Delta E$ to TARGET of at most 0.42 per application (dialogs round some values; maximum 3.4,
GIMP).

\paragraph{Strength curve.}
Every numeric field of the hidden program is set to $k$ times its distance from the field's neutral value
($k \in \{-1, 0, 0.25, 0.5, 0.75, 0.9, 1, 1.1, 1.25, 1.5, 2, 3\}$; hue fields are kept), rendered and scored
against TARGET, 1,200 renders per application.
Pass rates per application: $\times0.75$ darktable 26\%, RawTherapee 39\%, GIMP 30\%, Inkscape 35\%, Blender 42\%;
$\times1.25$ 34\%, 47\%, 34\%, 35\%, 48\%; $\times0.9$ and $\times1$ 100\% everywhere; $\times1.1$ 100\% except
GIMP 89\%, Blender 97\%, Inkscape 99\%, where every failing task has a field that the larger strength pushes past the
application's parameter bound, so the rendered program is not $\times1.1$.
$k \le 0.5$, $k \ge 1.5$ and the reversed program pass on no task.
Median closeness: reversed $-0.8$ to $-0.3$, $\times3$ $-0.2$ to $0.2$ per application.
The Sensitive check (Sec.~\ref{sec:checks}) skips a strength that runs into a parameter bound for the same reason.

\paragraph{Wrong edits.}
The checks already exclude tasks that doing nothing, a shift to the target's mean colour or a Reinhard colour transfer
would pass.
We add two wrong answers on all 500 tasks: the reversed program ($k=-1$, 0 passes), and a searched look-alike, the
program closest to TARGET found by a random search over only the library tools that are \emph{not} in the hidden
program (1 pass, an Inkscape task).

\paragraph{VLM judges.}
Candidates with a known answer are built from 75 tasks (15 per application): the hidden program (exact), rescaled to
$\times0.9$, $\times1.1$ (should pass), $\times0.5$, $\times1.5$ and with one step left out (should fail).
A rescaling that would push a parameter past the application's bound is replaced by a milder one, and candidates are
grouped by the strength actually rendered.
Each judge sees START, TARGET and the candidate and answers whether the candidate reaches the target's effect and
whether its strength is weaker, the same or stronger; the open judges judge each pair three times and the majority
is used, the closed judges (GPT-6.1, Opus~5.5) once.
Asked about the strength of the wrong candidates, the open judges (GLM-5.3, Qwen3.5-397B) answer ``the same as the
target'' for 41--79\% of them, the closed ones (GPT-6.1, Opus~5.5) for 17--45\%.
On real agent outputs with $\Delta E > 6$ to the target the open judges still accept 11--43\%, the closed ones 1--6\%.

\paragraph{Human study.}
Raters see START, TARGET and a candidate, never its origin or score, and answer whether the candidate reaches the
target's effect and, if not, whether it falls short, overshoots or does something else.
The 913 pairs (about 180 per application) consist of the hidden program (15), its rescalings to $\times0.5$--$1.5$
(598), programs with a step left out (59) and agent outputs (241); 33\% pass the metric.
Pairs are dealt in batches of about 150 that mix applications and candidate kinds.
So far one rater has judged all 913 pairs; further raters are being collected.
Humans pass 38\% of the pairs and 80\% of the hidden programs themselves; their pass rate rises smoothly with
closeness and crosses one half at 0.70--0.80 (30\% at 0.5--0.6, 51\% at 0.70--0.75, 67\% at 0.8--0.9).
Away from the boundary the metric agrees with them on 82\% (GPT-6.1 81\%, Opus~5.5 78\%, Qwen3.5 73\%, GLM-5.3
60\%); at $\times0.75$ and $\times1.25$ every scorer is near chance (48--58\%).
Per application the metric's agreement ranges over 67--80\%.

\subsection{Tasks}
\label{app:val-tasks}

\paragraph{Reachability.}
In the feedback condition the agent must use four rounds and is told the mean $\Delta E$ of each RESULT to TARGET;
GPT-6.1 then passes 45--88\% of the L2 tasks per application (darktable 51\%, GIMP 54\%, RawTherapee 73\%,
Inkscape 88\%, Blender 45\%).
The random search tries programs over the given tools, sees the distance to TARGET after every render and keeps
the best; within 64 renders it passes 8\% (darktable), 26\% (RawTherapee) and 17\% (GIMP) of the L2 tasks
and 3--6\% of the L3 tasks, within four renders 3--6\% of the L2 tasks.

\paragraph{Generated vs.\ real tasks.}
The 160 PlayRaw tasks are run like the generated ones on nine models (the Qwen3.5 sizes, GLM-5.3).
Mean pass rates, generated vs.\ real: L1 97\% vs.\ 97\%, L2 2.3\% vs.\ 2.5\%, L3 0.1\% vs.\ 0.8\%.
Spearman correlation of the model order between the two sets: by mean closeness 0.78 (L2) and 0.72 (L3),
by pass rate $\le 0.07$.

\paragraph{Tasks by generator.}
\label{app:gensplit}
Each generator wrote 250 of the 500 tasks. Every model's pass rate differs by at most 3 points between the two
halves at L2 (at most 4.4 at L3, GPT-6.1), and the generators show no advantage on their own tasks:
at L2, Qwen3.5-397B passes 9.9\% of its own tasks and 11.0\% of GLM-5.3's, and GLM-5.3 passes 5.6\% of its own and
2.8\% of Qwen3.5's, within the range of the other models (from 3.0 points more on the GLM-5.3 tasks to 2.0 more on
the Qwen3.5 tasks).

\subsection{Ranking}
\label{app:val-models}

\paragraph{Agents.}
OpenCUA 7B, 32B, 72B; UI-TARS 2B, 7B, 72B and UI-TARS-1.5-7B; GUI-Owl-1.5-32B; EvoCUA-32B; Holo4 27B and 35B.
Each runs in the OSWorld harness with 100 steps on the GUI tasks of all five applications, three seeds; the rank checks below use GIMP. The harness check runs OpenCUA-72B on 30 OSWorld-Verified tasks: 43\% (95\% interval [27, 60]) against the published
45.0\%. On GIMP L1 the five agents with a published OSWorld score are ordered the same way as by that score
(Spearman $\rho = 0.90$, $p = 0.04$).

\paragraph{Scale.}
GIMP L1, mean over three seeds: UI-TARS 1\% (2B, 7B), 30\% (72B); OpenCUA 60\% (7B), 70\% (32B), 68\% (72B; intervals
overlap). Qwen3.5 through the same interface, on GIMP: 0.3\% (0.8B), 1\% (2B), 18\% (4B), 47\% (9B), 72\% (27B), 59--61\%
(35B-A3B, 122B-A10B, 397B-A17B); averaged over the five applications, L2 is at most 1\% and L3 at most 0.2\% for every size.

\paragraph{Order and stability.}
L1 $\ge$ L2 $\ge$ L3 holds for 9 of the 11 agents; the two exceptions, UI-TARS 2B and 7B, are at 1--2\% on every
level.
Across seeds: Kendall $\tau$ of the L1 ranking 0.95--0.98, largest spread of a model's L1 pass rate 9 points,
$\tau$ of the L2 ranking by mean closeness 0.78--0.89.
On darktable every agent except Holo4 (27B 36\%, 35B 16\%; 95\% intervals 27--45\% and 8--24\%) passes at most 7\% at L1 in any seed.

\section{Results per application}
\label{app:results}

\subsection{Per application}
\label{app:per-app}

Tables~\ref{tab:app-prog} and~\ref{tab:app-gui} break the main results (Tab.~\ref{tab:main}) down
by application. Closeness per application and the per-seed values are in the released result files.

Inkscape is the easiest application: on programs, GPT-6.1 passes 65\% at L2 and 52\% at L3 there, against 4--11\% at L2 on the other
four, and every model does best on it. Its operations (filter primitives such as blur, colour matrices and drop shadows on
flat vector art) have a more direct visual reading than the parameters of a raw developer or a 3D renderer; the open
models still pass only 12\% at L2 and 4\% at L3 on it.

\begin{table*}[t]
  \centering
  \footnotesize
  \caption{\textbf{Program channel per application}: pass rate (\%), seeds as in Tab.~\ref{tab:main}. dt: darktable, RT: RawTherapee, Ink: Inkscape.}
  \label{tab:app-prog}
  \setlength{\tabcolsep}{3.4pt}
  \begin{tabular}{@{}l ccccc @{\hspace{7pt}} ccccc @{\hspace{7pt}} ccccc@{}}
    \toprule
    & \multicolumn{5}{c}{L1} & \multicolumn{5}{c}{L2} & \multicolumn{5}{c}{L3} \\
    \cmidrule(r{7pt}){2-6}\cmidrule(r{7pt}){7-11}\cmidrule{12-16}
    Model & dt & RT & GIMP & Ink & Blender & dt & RT & GIMP & Ink & Blender & dt & RT & GIMP & Ink & Blender \\
    \midrule
    Qwen3.5-0.8B           & 82 & 86 & 85 & 92 & 87 & 0 & 2 & 2 & 3 & 0 & 0 & 0 & 0 & 0 & 0 \\
    Qwen3.5-2B             & 92 & 93 & 90 & 91 & 94 & 1 & 1 & 0 & 2 & 1 & 0 & 0 & 0 & 1 & 0 \\
    Qwen3.5-4B             & 100 & 100 & 99 & 99 & 99 & 2 & 4 & 5 & 9 & 2 & 0 & 0 & 0 & 2 & 0 \\
    Qwen3.5-9B             & 100 & 100 & 100 & 100 & 100 & 2 & 5 & 4 & 14 & 1 & 0 & 0 & 0 & 4 & 0 \\
    Qwen3.5-27B            & 100 & 100 & 100 & 100 & 100 & 4 & 6 & 9 & 16 & 3 & 0 & 0 & 0 & 3 & 0 \\
    Qwen3.5-35B-A3B        & 100 & 100 & 100 & 100 & 100 & 2 & 5 & 6 & 16 & 2 & 0 & 0 & 0 & 4 & 0 \\
    Qwen3.5-122B-A10B      & 100 & 100 & 100 & 100 & 100 & 3 & 3 & 8 & 13 & 4 & 0 & 0 & 0 & 7 & 0 \\
    Qwen3.5-397B-A17B      & 100 & 100 & 100 & 100 & 100 & 5 & 12 & 6 & 24 & 5 & 0 & 0 & 0 & 10 & 1 \\
    \midrule
    GLM-5.3-Flash          & 100 & 100 & 100 & 100 & 100 & 0 & 5 & 1 & 9 & 6 & 0 & 0 & 0 & 2 & 0 \\
    \midrule
    DeepSeek-V4.1-Flash    & 100 & 100 & 100 & 100 & 100 & 2 & 4 & 4 & 14 & 3 & 0 & 1 & 0 & 11 & 1 \\
    \midrule
    Step-3.7-Flash         & 100 & 100 & 100 & 100 & 100 & 3 & 12 & 7 & 14 & 1 & 1 & 0 & 0 & 2 & 0 \\
    \midrule
    Nemotron-3.5-Super-VL  & 100 & 100 & 100 & 100 & 100 & 4 & 4 & 9 & 12 & 5 & 0 & 0 & 0 & 2 & 1 \\
    \midrule
    Claude Opus 5.5        & 100 & 100 & 100 & 100 & 100 & 4 & 11 & 10 & 34 & 8 & 1 & 2 & 0 & 28 & 3 \\
    \midrule
    GPT-6.1                & 100 & 100 & 100 & 100 & 100 & 4 & 9 & 11 & 65 & 6 & 5 & 2 & 3 & 52 & 5 \\
    \bottomrule
  \end{tabular}
\end{table*}

\begin{table*}[t]
  \centering
  \footnotesize
  \caption{\textbf{GUI channel per application}: pass rate (\%), 100 steps, seeds as in Tab.~\ref{tab:main}. Tasks per level: darktable 53, RawTherapee 100, GIMP 100, Inkscape 73, Blender 100.}
  \label{tab:app-gui}
  \setlength{\tabcolsep}{3.4pt}
  \begin{tabular}{@{}l ccccc @{\hspace{7pt}} ccccc @{\hspace{7pt}} ccccc@{}}
    \toprule
    & \multicolumn{5}{c}{L1} & \multicolumn{5}{c}{L2} & \multicolumn{5}{c}{L3} \\
    \cmidrule(r{7pt}){2-6}\cmidrule(r{7pt}){7-11}\cmidrule{12-16}
    Model & dt & RT & GIMP & Ink & Blender & dt & RT & GIMP & Ink & Blender & dt & RT & GIMP & Ink & Blender \\
    \midrule
    Qwen3.5-0.8B           & 0 & 0 & 0 & 0 & 2 & 0 & 0 & 0 & 0 & 0 & 0 & 0 & 0 & 0 & 0 \\
    Qwen3.5-2B             & 0 & 0 & 1 & 0 & 2 & 0 & 0 & 0 & 0 & 0 & 0 & 0 & 0 & 0 & 0 \\
    Qwen3.5-4B             & 0 & 0 & 18 & 2 & 0 & 0 & 0 & 1 & 0 & 2 & 0 & 0 & 0 & 0 & 0 \\
    Qwen3.5-9B             & 0 & 0 & 47 & 3 & 0 & 0 & 0 & 1 & 0 & 2 & 0 & 0 & 0 & 0 & 0 \\
    Qwen3.5-27B            & 1 & 4 & 72 & 2 & 3 & 0 & 0 & 3 & 0 & 0 & 0 & 0 & 1 & 0 & 0 \\
    Qwen3.5-35B-A3B        & 1 & 3 & 61 & 1 & 14 & 0 & 0 & 1 & 0 & 1 & 0 & 0 & 0 & 0 & 0 \\
    Qwen3.5-122B-A10B      & 0 & 2 & 61 & 3 & 11 & 0 & 0 & 3 & 0 & 1 & 0 & 0 & 0 & 0 & 0 \\
    Qwen3.5-397B-A17B      & 4 & 1 & 59 & 3 & 5 & 0 & 0 & 3 & 0 & 1 & 0 & 0 & 0 & 0 & 0 \\
    \midrule
    GLM-5.3-Flash          & 81 & 70 & 51 & 49 & 49 & 0 & 5 & 4 & 12 & 2 & 0 & 0 & 0 & 8 & 0 \\
    \midrule
    DeepSeek-V4.1-Flash    & 0 & 7 & 9 & 1 & 23 & 0 & 2 & 1 & 0 & 3 & 0 & 0 & 2 & 0 & 0 \\
    \midrule
    Step-3.7-Flash         & 0 & 0 & 0 & 0 & 0 & 0 & 0 & 0 & 0 & 0 & 0 & 0 & 0 & 0 & 0 \\
    \midrule
    Nemotron-3.5-Super-VL  & 0 & 0 & 0 & 0 & 0 & 0 & 0 & 0 & 0 & 0 & 0 & 0 & 0 & 0 & 0 \\
    \midrule
    UI-TARS-2B             & 0 & 0 & 1 & 0 & 1 & 0 & 0 & 2 & 0 & 1 & 0 & 0 & 0 & 0 & 0 \\
    UI-TARS-7B             & 0 & 0 & 1 & 0 & 1 & 0 & 0 & 2 & 0 & 2 & 0 & 0 & 0 & 0 & 0 \\
    UI-TARS-72B            & 0 & 2 & 30 & 1 & 1 & 0 & 0 & 2 & 0 & 2 & 0 & 0 & 0 & 0 & 0 \\
    UI-TARS-1.5-7B         & 1 & 0 & 46 & 2 & 0 & 0 & 0 & 2 & 1 & 2 & 0 & 0 & 0 & 0 & 0 \\
    \midrule
    OpenCUA-7B             & 1 & 1 & 60 & 0 & 0 & 0 & 0 & 1 & 0 & 0 & 0 & 0 & 0 & 0 & 0 \\
    OpenCUA-32B            & 1 & 1 & 70 & 1 & 2 & 0 & 0 & 1 & 0 & 1 & 0 & 0 & 0 & 0 & 0 \\
    OpenCUA-72B            & 3 & 2 & 68 & 2 & 7 & 0 & 0 & 2 & 0 & 2 & 0 & 0 & 0 & 0 & 0 \\
    \midrule
    GUI-Owl-1.5-32B        & 0 & 2 & 58 & 1 & 0 & 0 & 0 & 2 & 0 & 1 & 0 & 0 & 0 & 0 & 0 \\
    \midrule
    EvoCUA-32B             & 1 & 4 & 80 & 2 & 0 & 0 & 1 & 1 & 0 & 2 & 0 & 0 & 0 & 0 & 0 \\
    \midrule
    Holo4-35B              & 16 & 24 & 98 & 24 & 22 & 1 & 1 & 3 & 3 & 1 & 0 & 0 & 0 & 1 & 0 \\
    Holo4-27B              & 36 & 48 & 99 & 26 & 54 & 1 & 1 & 2 & 5 & 2 & 0 & 0 & 0 & 9 & 0 \\
    \midrule
    Claude Opus 5.5        & 100 & 100 & 100 & 96 & 98 & 0 & 10 & 12 & 42 & 4 & 0 & 2 & 3 & 34 & 4 \\
    \midrule
    GPT-6.1                & 96 & 100 & 84 & 75 & 92 & 17 & 25 & 30 & 64 & 27 & 11 & 11 & 5 & 59 & 3 \\
    \bottomrule
  \end{tabular}
\end{table*}

\begin{figure*}[!tp]
  \centering
  \begin{subfigure}[b]{0.24\textwidth}\centering
    \includegraphics[width=\linewidth]{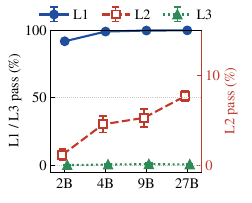}\caption{Qwen3.5, program}\label{fig:scaling-prog}
  \end{subfigure}\hfill
  \begin{subfigure}[b]{0.24\textwidth}\centering
    \includegraphics[width=\linewidth]{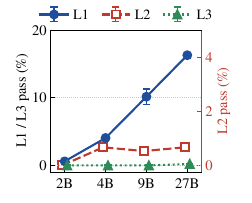}\caption{Qwen3.5, GUI}\label{fig:scaling-qwen}
  \end{subfigure}\hfill
  \begin{subfigure}[b]{0.24\textwidth}\centering
    \includegraphics[width=\linewidth]{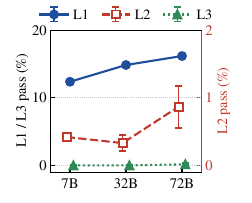}\caption{OpenCUA, GUI}\label{fig:scaling-opencua}
  \end{subfigure}\hfill
  \begin{subfigure}[b]{0.24\textwidth}\centering
    \includegraphics[width=\linewidth]{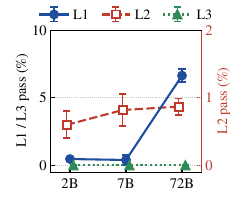}\caption{UI-TARS, GUI}\label{fig:scaling-uitars}
  \end{subfigure}
  \caption{\textbf{Size raises L1 far more than L2 / L3.} Pass rate per level against model size, mean over the five applications
  (each weighs the same); error bars: one standard deviation over seeds (GUI: 3 seeds, 100 steps; program: 5 seeds).
  L1 and L3 use the left axis, L2 the right axis (red), whose range is much smaller. Episodes in which the agent left the
  application hung count as failures.}
  \label{fig:scaling}
\end{figure*}

\paragraph{GPT-6.1's GUI failures at L1.}
They are operational: in GIMP it often ends with a reply that OSWorld's parser rejects and then applies the edit a second
time (13 of its 16 failures), and in Inkscape it gets stuck in filter dialogs until the 100 steps run out (23 of 73
episodes; Opus 5.5 none). Limited to one action per step it gains on Inkscape (75\% $\to$ 90\%) but loses on GIMP
(84\% $\to$ 30\%), where multi-field dialogs break up.

\paragraph{Confidence intervals.}
\label{app:ci}
Resampling the tasks of each application with replacement (2,000 draws; seeds averaged per task, the five applications
weighing the same, as in Tab.~\ref{tab:main}), the 95\% interval of every pass rate in Tab.~\ref{tab:main} has a
half-width of at most 1.8 (L1), 2.9 (L2) and 2.4 (L3) points on programs and 4.6, 4.2 and 3.3 points on the GUI.
The closed models' GUI L2 rates, for example, are 13.7\% [10.8, 16.6] (Opus 5.5) and 32.7\% [28.4, 36.8] (GPT-6.1).

\subsection{Error after each round}
\label{app:rounds}

\begin{figure}[!htbp]
  \centering
  \includegraphics[width=\linewidth]{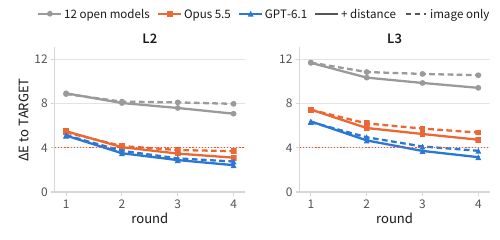}
  \caption{\textbf{Error after each round} when the agent must use four rounds, with the RESULT image only (dashed)
  or also its measured distance to TARGET (solid). Mean $\Delta E$ over the episodes present in both runs; dotted:
  the pass bound $\Delta E = 4$.}
  \label{fig:rounds}
\end{figure}

Fig.~\ref{fig:rounds} shows, for the forced four-round runs of Sec.~\ref{sec:why}, how far the result is from
the target after each round.

\subsection{Other pass rules}
\label{app:thresholds}
The pass rule (closeness $\ge 0.75$ and $\Delta E \le 4$) sits between two requirements: most of the edit done, and a
small remaining error. Tab.~\ref{tab:thresholds} rescores every episode under looser and stricter rules (legality
gates kept; same averaging as Tab.~\ref{tab:main}).
L1 hardly moves: on programs every rule gives 97.7--98.3\% on average, and on the GUI the best open model stays at
57--65\% and the closed models at 87--99\%; the model order at L1 is unchanged (Kendall $\tau \ge 0.93$ against the
default rule).
L2 and L3 rise as the rule loosens, but the cliff remains under every rule: with $\Delta E \le 4$ alone, which also
passes a third of the $\times0.5$ and $\times1.5$ programs (Sec.~\ref{sec:cliff}), the best model passes 46\% of L2 on
programs and 61\% on the GUI, against 100\% and 99\% at L1, and open models at most 32\% and 14\%.
The closed models lead and GPT-6.1 is first under every rule; the order of all models changes little between 0.7 and
0.8 ($\tau \ge 0.87$ at L2) and more at the extremes, where most models are near 0 (closeness $\ge 0.9$) or the rule no
longer separates wrong amounts ($\Delta E \le 4$ alone).
Per model, the leaders at L2 keep their places under every rule (GPT-6.1, Opus 5.5 and Qwen3.5-397B-A17B on programs;
GPT-6.1, Opus 5.5, GLM-5.3 and Holo4-27B on the GUI); the lower $\tau$ comes from swaps among models whose pass rates
under the default rule differ by less than 3.5 points.

\begin{table*}[!htbp]
  \centering
  \small
  \caption{\textbf{Other pass rules.} Program: mean over all models at L1; best open / best closed model at L2 and L3.
  GUI: best open / best closed model at every level. $\tau$: Kendall's $\tau$ of the order of all models at L2 against
  the default rule (bold).}
  \label{tab:thresholds}
  \setlength{\tabcolsep}{4pt}
  \begin{tabular}{@{}l c cc c@{\hspace{14pt}}ccc c@{}}
    \toprule
    & \multicolumn{4}{c}{Program channel} & \multicolumn{4}{c}{GUI channel} \\
    \cmidrule(r{14pt}){2-5}\cmidrule{6-9}
    Pass rule & L1 & L2 & L3 & $\tau$ (L2) & L1 & L2 & L3 & $\tau$ (L2) \\
    \midrule
    closeness $\ge 0.60$, $\Delta E \le 4$ & 98.2 & 22.2 / 33.6 & 5.3 / 22.2 & 0.82 & 62.7 / 99.1 & 10.0 / 50.9 & 3.5 / 28.6 & 0.83 \\
    closeness $\ge 0.70$, $\Delta E \le 4$ & 98.1 & 14.5 / 24.2 & 3.2 / 15.8 & 0.87 & 61.0 / 99.1 & 6.7 / 39.5 & 2.5 / 21.8 & 0.92 \\
    \textbf{closeness $\ge$ 0.75, $\Delta E \le 4$} & \textbf{98.0} & \textbf{10.5 / 19.0} & \textbf{2.6 / 13.4} & -- & \textbf{60.1 / 98.8} & \textbf{4.7 / 32.7} & \textbf{2.0 / 17.8} & -- \\
    closeness $\ge 0.80$, $\Delta E \le 4$ & 97.9 & 6.7 / 15.6 & 2.0 / 10.8 & 0.91 & 58.8 / 98.8 & 3.6 / 26.0 & 1.4 / 14.6 & 0.92 \\
    closeness $\ge 0.90$, $\Delta E \le 4$ & 97.7 & 1.4 / 8.8 & 1.0 / 5.8 & 0.64 & 56.7 / 95.8 & 1.1 / 11.8 & 0.5 / 6.9 & 0.79 \\
    $\Delta E \le 4$ alone & 98.3 & 31.9 / 45.6 & 10.6 / 33.2 & 0.78 & 64.5 / 99.3 & 14.1 / 60.7 & 6.6 / 41.5 & 0.73 \\
    \bottomrule
  \end{tabular}
\end{table*}

\subsection{Interventions}
\label{app:interventions}

Tab.~\ref{tab:interventions-full} gives the pass rates before and after each intervention of Sec.~\ref{sec:why}, the
change of the median closeness and the significance of both.

\begin{table*}[!tp]
  \centering
  \caption{\textbf{Interventions in full} (Tab.~\ref{tab:interventions}). Each condition against the default run on the same L2 episodes: pass rate
  before $\to$ after (\%) and change of the median closeness ($\Delta$cl).
  Open: the twelve open models, in the knowledge and TARGET$-$START rows on 30-task subsets per application (1,800 pairs),
  in the round rows on all 500 tasks (6,000 pairs).
  Opus 5.5 / GPT-6.1: knowledge and TARGET$-$START rows on the same 30-task subsets (program 150 pairs, GUI 125), rounds on all 500 tasks.
  --: not run. $^{*}p<0.001$, $^{\dagger}p<0.01$, $^{\ddagger}p<0.05$ (McNemar for pass, Wilcoxon for closeness).}
  \label{tab:interventions-full}
  \setlength{\tabcolsep}{4pt}
  \resizebox{\textwidth}{!}{%
  \begin{tabular}{@{}l cc@{\hspace{10pt}}cc@{\hspace{10pt}}cc@{}}
    \toprule
    & \multicolumn{2}{c}{Open} & \multicolumn{2}{c}{Opus 5.5} & \multicolumn{2}{c}{GPT-6.1} \\
    \cmidrule(r{10pt}){2-3}\cmidrule(r{10pt}){4-5}\cmidrule{6-7}
    Condition & Pass (\%) & $\Delta$cl & Pass (\%) & $\Delta$cl & Pass (\%) & $\Delta$cl \\
    \midrule
    \multicolumn{7}{@{}l}{\emph{Knowledge about the tools}} \\
    The application's full manual               & 5.7 $\to$ 4.7  & 0.00           & 12.7 $\to$ 14.0 & 0.00 & 17.3 $\to$ 16.7 & 0.00 \\
    A reference card per tool                   & 5.7 $\to$ 6.3  & $-$0.07$^{*}$ & 12.7 $\to$ 16.0 & $+$0.02 & 17.3 $\to$ 24.0$^{\ddagger}$ & $+$0.04$^{*}$ \\
    Cards + each tool's effect on an image      & 5.7 $\to$ 6.8  & $-$0.12$^{*}$ & 12.7 $\to$ 13.3 & $+$0.02$^{\ddagger}$ & 17.3 $\to$ 25.3$^{\ddagger}$ & $+$0.04$^{*}$ \\
    Each setting's effect measured on START     & 5.7 $\to$ 4.3$^{\ddagger}$  & $-$0.19$^{*}$  & 12.7 $\to$ 14.0 & $+$0.02 & 17.3 $\to$ 25.3$^{\ddagger}$ & $+$0.05$^{*}$ \\
    \midrule
    \multicolumn{7}{@{}l}{\emph{The measured difference}} \\
    Statistics of TARGET $-$ START              & 5.7 $\to$ 14.9$^{*}$ & $+$0.03$^{\ddagger}$ & 12.7 $\to$ 34.0$^{*}$ & $+$0.15$^{*}$ & 17.3 $\to$ 47.3$^{*}$ & $+$0.17$^{*}$ \\
    TARGET $-$ START as an image                & 5.7 $\to$ 6.9  & $-$0.01$^{*}$ & -- & -- & -- & -- \\
    Statistics of TARGET $-$ START, GUI         & -- & -- & 10.4 $\to$ 30.4$^{*}$ & $+$0.10$^{*}$ & 28.0 $\to$ 39.2$^{\ddagger}$ & $+$0.07$^{*}$ \\
    \midrule
    \multicolumn{7}{@{}l}{\emph{Looking again}} \\
    Four forced rounds, each RESULT shown       & 5.6 $\to$ 8.2$^{*}$  & $+$0.04$^{*}$ & 13.4 $\to$ 29.8$^{*}$ & $+$0.10$^{*}$ & 19.0 $\to$ 50.4$^{*}$ & $+$0.19$^{*}$ \\
    \quad + $\Delta E$ to TARGET each round      & 5.6 $\to$ 13.2$^{*}$ & $+$0.11$^{*}$ & 13.4 $\to$ 38.0$^{*}$ & $+$0.18$^{*}$ & 19.0 $\to$ 62.2$^{*}$ & $+$0.21$^{*}$ \\
    \bottomrule
  \end{tabular}}
\end{table*}

\section{Limitations}
\label{app:limitations}

\paragraph{Visual results only.}
The metric compares the re-rendered result with TARGET in colour space, so \bench applies to software whose output
is an image or a render. Effect judgement exists elsewhere too (how loud, how smooth, how much), but there the result
is not seen and needs another reference metric. The findings concern visual effects, which is what the five
applications produce.

\paragraph{Global, single-image edits.}
The generated tasks use global operations on one image; local edits, masks and multi-image projects are not covered.
These add localisation on top of the same judgement of amount, so they are unlikely to make the tasks easier: agents
that already miss the amount of a global edit must still judge it on a local one.

\paragraph{Two closed models.}
The closed models are the two available to us. They serve to show that the tasks are solvable and that the cliff is
not specific to open models; every finding is also shown on the twelve open models.

\paragraph{Human agreement.}
Human judgements agree with the metric on 72.5\% of the pairs but are near chance at the boundary
($\times0.75$, $\times1.25$), where no scorer exceeds 58\%; the threshold matches the human boundary (0.71, CI
0.60--0.83) only on average, the study has a single rater so far, who is not a professional colorist.

\end{document}